\documentclass[11pt]{article}

\usepackage[final]{acl}

\usepackage{times}
\usepackage{latexsym}

\usepackage[T1]{fontenc}

\usepackage[utf8]{inputenc}
\usepackage{amsmath}
\usepackage{subcaption}
\usepackage{microtype}
\usepackage{lineno}
\usepackage{inconsolata}

\usepackage{graphicx}
\usepackage{xcolor}
\usepackage{listings}
\usepackage{amssymb}
\usepackage{algorithm}
\usepackage{algpseudocode}
\usepackage{booktabs}

\usepackage{multirow}

\title{SMADE-IE: Sparse Multi-Agent Framework with Evidence-Driven Debate for Zero-Shot Information Extraction}

\author{
Kenfeng Huang$^{\clubsuit}$,
Yi Cai$^{\clubsuit}$,
Xin Wu$^{\clubsuit}$,
Zikun Deng$^{\clubsuit}$,
Li Yuan$^{\heartsuit\dagger}$\\
$^{\heartsuit}$School of Artificial Intelligence, South China Normal University,
Foshan, China\\
$^{\clubsuit}$School of Software Engineering,
South China University of Technology,
Guangzhou, China\\
\texttt{kenfenghuang@gmail.com}, \texttt{\{ycai, xinwu, zkdeng\}@scut.edu.cn}\\
\texttt{yuanli@scnu.edu.cn}
}

\begin{document}
\maketitle
\let\thefootnote\relax
\footnotetext{$^{\dagger}$ Corresponding author.}

\begin{abstract}
Zero-shot information extraction (IE) with large language models (LLMs) enables adaptation to new schemas and domains without task-specific training. Existing methods mainly follow three paradigms. Monolithic prompting is efficient but prone to missed mentions, boundary errors, and type confusion. Each-type prompting improves type-level focus but may produce overlapping or conflicting predictions, while multi-agent debate can resolve such conflicts at the cost of irrelevant context, redundant interactions, and high token overhead. To address these issues, we propose \textsc{SMADE-IE}, a sparse and evidence-driven multi-agent framework. An Adaptive Mode Selector routes simple inputs to a lightweight Global Extraction Mode and ambiguous inputs to a Type-Centric Extraction Mode based on sample complexity and relevant types. Cross-type conflicts are resolved by an Evidence-Driven Debate module that uses Toulmin-style arguments, external evidence scoring, Beta-based confidence updates, and early stopping. Experiments on nine benchmarks covering NER, RE, and JERE show that \textsc{SMADE-IE} improves average Partial F1 over the strongest baselines by 11.37, 3.83, and 14.46 points, respectively. Compared with the multi-agent baseline \textsc{CrossAgentIE}, \textsc{SMADE-IE} reduces token consumption by 85.1\% on DocRED and 80.1\% on CrossRE, demonstrating substantially higher inference efficiency. Code is available at \url{https://github.com/Cppys/SMADE-IE}.
\end{abstract}

\section{Introduction}

Information Extraction (IE) aims to extract structured facts from unstructured text~\citep{li-etal-2023-extracting}, supporting applications such as knowledge graph construction and question answering \citep{liu2020event,jain2020domain,zhang2025gkg,yuan2026hybrid}. Representative tasks include named entity recognition (NER), relation extraction (RE), and joint entity and relation extraction (JERE) \citep{zeng-etal-2014-relation,zheng-etal-2017-joint}. Although supervised and instruction-tuned IE methods achieve strong performance~\citep{lu-etal-2022-unified,lou2023universalinformationextractionunified}, they often rely on extensive annotations, task-specific tuning, or fixed schemas, limiting adaptability to evolving entity types, relation patterns, and domains~\citep{sainz2024gollie,fuente-etal-2025-guidex}.

\begin{figure}[t]
    \centering
    \includegraphics[scale=.43]{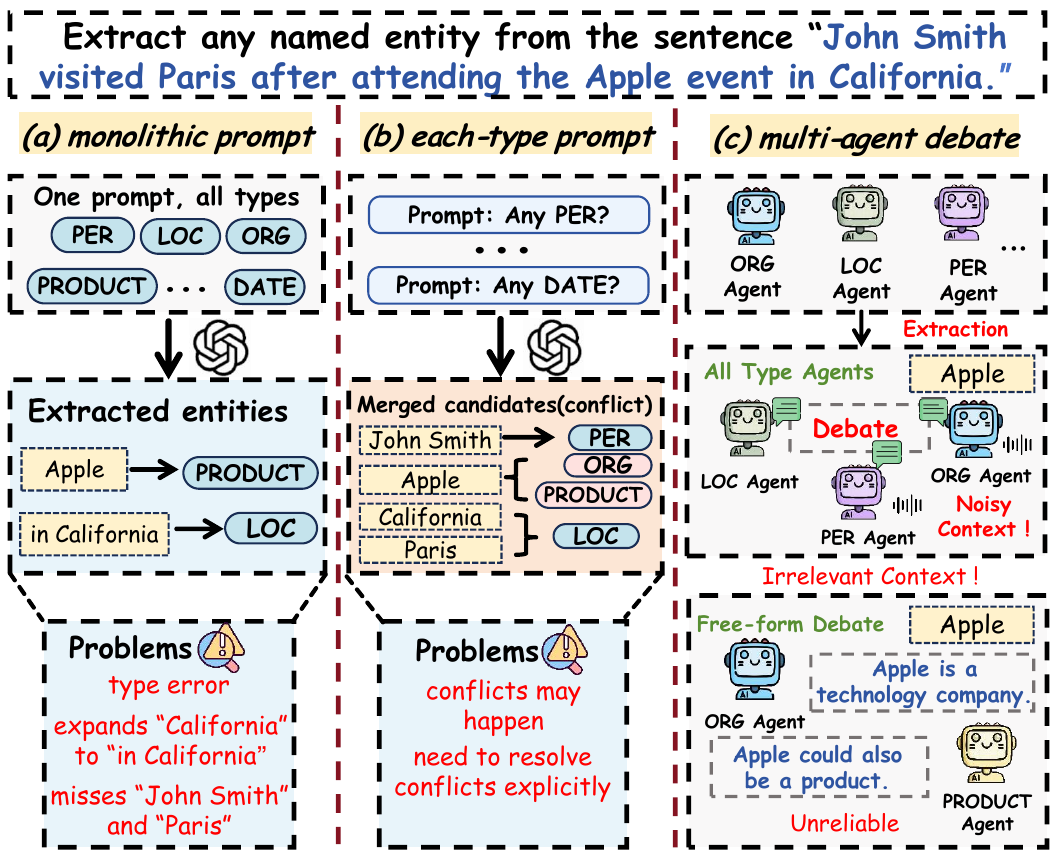}
    \caption{Existing LLM-based zero-shot IE methods.}
    \vspace{-4mm}
    \label{fig:intro_example}
\end{figure}

Some studies leverage the extensive background knowledge of LLMs for zero-shot IE, enabling flexible adaptation to new schemas and domains~\citep{xu2024largelanguagemodelsgenerative}. Early LLM-based zero-shot IE methods typically use a \emph{monolithic-prompt} strategy, extracting all elements in a single stage~\citep{xie-etal-2023-empirical,li2024gno}. Despite its efficiency, this strategy requires jointly predicting entity spans and types, which often leads to missed mentions, boundary errors, and type confusion. As shown in Figure~\ref{fig:intro_example}(a), the model misses ``John Smith'', expands ``California'' to ``in California'', and misclassifies ``Apple'' as \textsc{Product}. To alleviate these issues, recent work has introduced \emph{each-type prompting}~\citep{li-etal-2023-revisiting-large,li2024gno}, which extracts each entity or relation type independently using separate prompts. However, independently generated type-specific predictions can overlap or conflict with one another. For example, in Figure~\ref{fig:intro_example}(b), ``Apple'' is simultaneously assigned the types \textsc{Organization} and \textsc{Product}.

To mitigate such conflicts, recent \emph{multi-agent debate} methods employ specialized agents for different entity or relation types and use inter-agent discussions to resolve conflicting predictions~\citep{guan-etal-2025-mmd,crossagentie,situ2026grounded,situ2026self}. However, exhaustive agent interactions still limit their efficiency and reliability~\citep{yan2026tide}. First, since each sample typically contains only a small subset of all entity or relation types, as shown in Table~\ref{tab:intro-sparsity}, selecting all types for every sample introduces unnecessary token overhead and computational cost. Second, incorrect candidate types inject irrelevant context into the debate, distracting agents from meaningful evidence and leading to unreliable decisions~\citep{Fan_Yoon_Ji_2026,pmlr-v202-shi23a,liu-etal-2024-lost,yang2025llmreasoningdistractedirrelevant,tian2026multiagent,shen-etal-2026-truth}. For instance, as shown in Figure~\ref{fig:intro_example}(c), three agents (\emph{PER}, \emph{ORG}, and \emph{LOC}) label the same entity with different types, causing redundant discussions that obscure useful evidence and increase the risk of incorrect adjudication.
Finally, free-form debates lack explicit structures for modeling claims, points of contention, and supporting evidence, making it difficult to systematically aggregate confidence scores and reach stable decisions~\citep{choi2025debate}. The same example further illustrates this issue: the \textsc{Product}
agent argues that ``Apple could also be a product,'' whereas the
\textsc{Organization} agent claims that ``Apple is a technology company.''
Without a structured mechanism to distinguish reliable evidence from
noisy or irrelevant arguments, such debates may fail to converge and
incur substantial token overhead~\citep{choi2025debate}. Beyond
cross-type conflicts, the JERE task further requires consistency between
entity and relation predictions, since relation constraints may reveal
missing entities or incompatible entity types. This motivates an explicit entity--relation alignment mechanism for joint extraction.

\begin{figure*}[t]
  \centering
  \includegraphics[scale=.48]{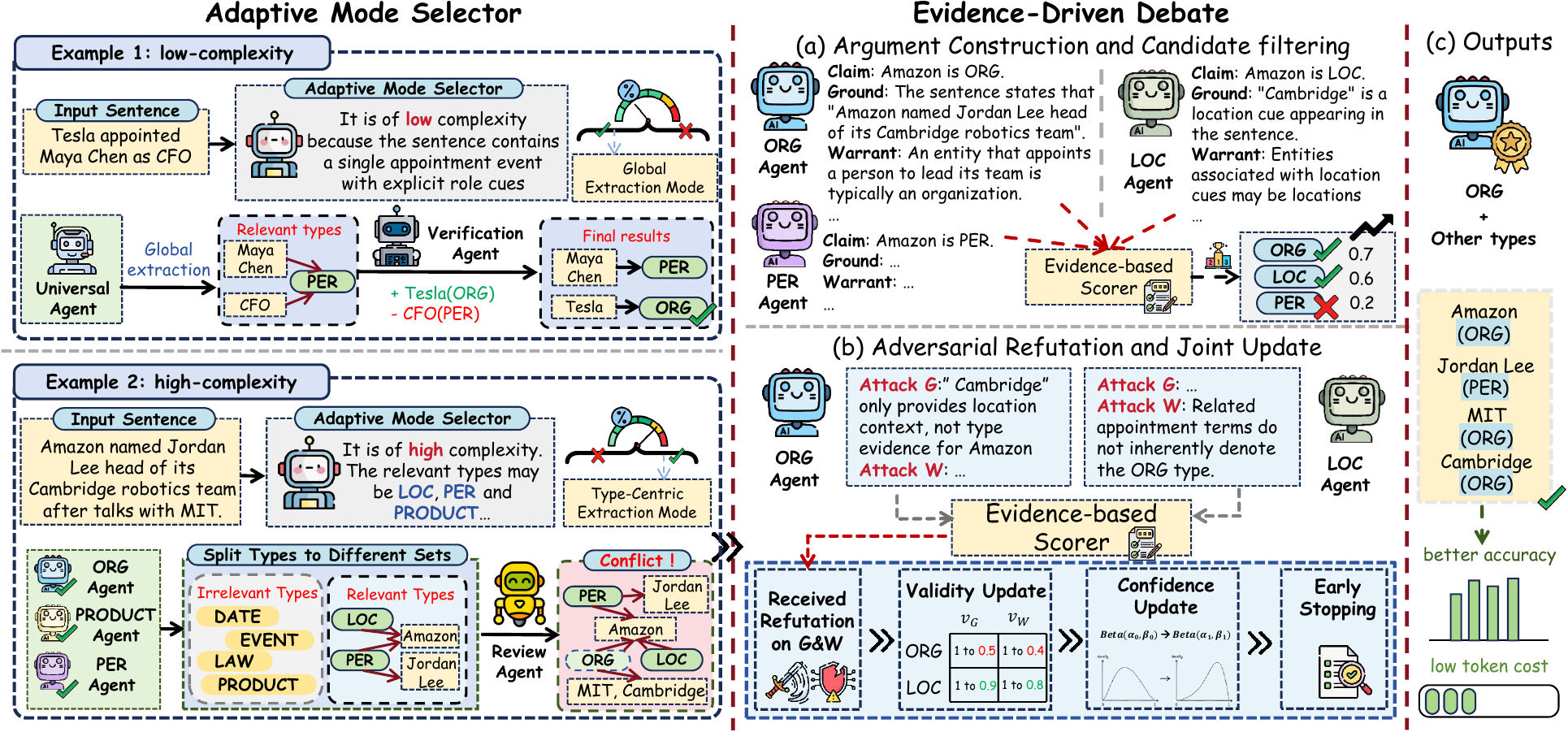}
  \caption{
  Overview of \textsc{SMADE-IE}. The Router Agent selects extraction modes. For conflicted predictions, Evidence-Driven Debate performs Toulmin-based ranking, Bayesian debate, and conflict resolution.
  }
  \label{fig:overview}
  \vspace{-4mm}
\end{figure*}

\begin{table}[t]
\centering
\small
\setlength{\tabcolsep}{3.5pt}
\begin{tabular}{llcc}
\toprule
\textbf{Task} & \textbf{Dataset} & \textbf{Ontology Size} & \textbf{Avg. Types / Sample} \\
\midrule
\multirow{2}{*}{NER} & OntoNotes5 & 18 & 0.99 \\
                     & CrossRE    & 38 & 2.17 \\
\midrule
\multirow{3}{*}{RE}  & DocRED     & 32 & 1.60 \\
                     & REDFM      & 32 & 1.95 \\
                     & SciERC    & 8 & 1.28 \\
\bottomrule
\end{tabular}
\caption{Statistics of some zero-shot IE benchmarks.}
\label{tab:intro-sparsity}
\vspace{-5mm}
\end{table}
To address these challenges, we propose \textsc{SMADE-IE}, a sparse and evidence-driven multi-agent framework for zero-shot IE. As shown in Figure~\ref{fig:overview}, \textsc{SMADE-IE} balances extraction reliability and computational cost through two components. First, the Adaptive Mode Selector estimates sample complexity and selects only relevant types: simple inputs follow the lightweight Global Extraction Mode, while complex inputs enter the Type-Centric Extraction Mode, reducing unnecessary token overhead and irrelevant reasoning noise. Second, for type conflicts in the Type-Centric Extraction Mode, \textsc{SMADE-IE} introduces a structured debate mechanism. 
Unlike free-form debate, it decomposes candidate predictions into Toulmin-style argumentative components~\citep{gupta-etal-2024-harnessing}, allowing agents to attack the evidence and reasoning behind each claim. 
An external evidence scorer and Bayesian Beta updates then aggregate support and refutation signals into evidence-centered confidence estimates for final adjudication.
The main contributions are as follows:
\begin{itemize}
    \item We present \textsc{SMADE-IE}, a sparse multi-agent framework for zero-shot information extraction. With the Adaptive Mode Selector, it automatically dispatches samples to either the Global Extraction Mode or the Type-Centric Extraction Mode based on their complexity, balancing efficiency and performance.

    \item We introduce the Evidence-Driven Debate mechanism that structures each conflict into Toulmin-style components, and combines external evidence scoring with Bayesian Beta updates for more reliable multi-agent conflict adjudication.

    \item We evaluate \textsc{SMADE-IE} on 9 zero-shot IE datasets and show its advantages in both extraction performance and token efficiency.
\end{itemize}

\section{Methodology}
\label{sec:method}
In this section, we first introduce the definition of zero-shot IE tasks and then describe \textsc{SMADE-IE} in detail. As shown in Figure~\ref{fig:overview}, the Adaptive Mode Selector dispatches each input to either the lightweight Global Extraction Mode or the finer-grained Type-Centric Extraction Mode. For cross-type conflicts in the Type-Centric Extraction Mode, we introduce the Evidence-Driven Debate module for structured conflict adjudication. For JERE, an Iterative Entity--Relation Alignment step further enforces ontology consistency between entity and relation predictions.

\subsection{Problem Formulation}
\label{sec:formulation}

Given a sentence $s$ and predefined sets of entity types $\mathcal{T}_e$ and relation types $\mathcal{T}_r$, NER aims to identify all entity mentions $\{e_1, \dots, e_n\}$ in $s$ and assign a type $t_e^i \in \mathcal{T}_e$ to each, producing a set of typed entities $E = \{ (e_i, t_e^i) \}$. RE requires first identifying the entity mentions $\{e_1, \dots, e_n\}$ and then predicting, for each ordered pair of entities $(e_i, e_j)$, the relation $r_{ij} \in \mathcal{T}_r$ that holds between them, producing a set of relations $R = \{ (e_i, r_{ij}, e_j) \mid r_{ij} \in \mathcal{T}_r \}$. Finally, JERE simultaneously recognizes all entities and relations, yielding a set of fully typed quintuples $R^* = \{ (e_i, t_e^i, r_{ij}, e_j, t_e^j) \}$, where each quintuple specifies a head entity, a tail entity, their types, and the relation connecting them. Since zero-shot NER and RE are handled by the same LLM framework in \textsc{SMADE-IE} and differ only in the task-definition prompt, we present \textsc{SMADE-IE} from the zero-shot NER perspective.

\subsection{Adaptive Mode Selector}
\label{sec:mode-selection}

\textsc{SMADE-IE} dynamically selects the inference extraction mode for each input, reducing redundant calls while preserving type-level scrutiny in ambiguous cases. Inspired by the fast-and-slow thinking paradigm in LLM reasoning~\citep{xu2023mind,li202512surveyreasoning}, we introduce a Router Agent to estimate both the relevant type subset and the sample complexity. Given a sentence $s$ and the entity type set $\mathcal{T}_e$, the Router Agent produces,
\begin{equation}
(\mathcal{A}, c) = \mathrm{LLM}(s, P^{sel}_{\mathcal{T}_e}),
\label{eq:selector}
\end{equation}
where $\mathrm{LLM}(\cdot)$ denotes the large language model, $P^{sel}_{\mathcal{T}_e}$ is the selector prompt, $\mathcal{A} \subseteq \mathcal{T}_e$ is the candidate type subset, and $c \in \{\mathrm{low}, \mathrm{med}, \mathrm{high}\}$ denotes the estimated sample complexity. We adopt three complexity levels instead of binary labels to better capture borderline uncertainty~\citep{mahajan2026mindgapelicitationprotocols}. The estimates are then mapped into two execution modes: low-complexity inputs use the lightweight Global Extraction Mode, while medium- and high-complexity inputs use the Type-Centric Extraction Mode.

\subsection{Global Extraction Mode}
\label{sec:global-mode}

To effectively balance extraction performance and computational cost, particularly in terms of token usage, \textsc{SMADE-IE} adopts the Global Extraction Mode for low-complexity inputs ($c = \mathrm{low}$). It employs two sequential LLM agents: a Universal Agent that first generates a one-shot candidate set, followed by a Verification Agent that performs targeted verification and refinement.

\subsubsection{Universal Agent}
For simple samples, we first apply the monolithic-prompt method~\citep{xie-etal-2023-empirical} to extract all candidate entities with types in $\mathcal{T}_e$. Formally, given the global extraction prompt $P_{\mathcal{T}_e}^{\mathrm{global}}$, the output of the Universal Agent is defined as
\begin{equation}
E_U = \mathrm{LLM}(s, P_{\mathcal{T}_e}^{\mathrm{global}}),
\label{eq:universal}
\end{equation}
where every element $(e_i, t^i_e) \in E_U$ pairs a candidate entity with a type.

\subsubsection{Verification Agent}

Although the Universal Agent extracts all elements in a single pass, its coarse-grained generation strategy may still miss valid entities or introduce redundant candidates. To improve extraction reliability, the Verification Agent, inspired by the clean-up module in G\&O~\citep{li2024gno}, performs bidirectional refinement over the candidate set $E_U$. Given the verification prompt $P_{\mathcal{T}_e}^{\mathrm{ver}}$ and $E_U$, the Verification Agent inserts missing entities and removes unsupported candidates:
\begin{equation}
(E^{+}, E^{-}) = \mathrm{LLM}(s, E_U, P_{\mathcal{T}_e}^{\mathrm{ver}}),
\label{eq:verify}
\end{equation}
where $E^{+}$ and $E^{-}$ denote the sets of entities to be inserted and deleted, respectively. Accordingly, the final output of the Global Extraction Mode is:
\begin{equation}
E = (E_U \setminus E^{-}) \cup E^{+}.
\end{equation}

\subsection{Type-Centric Extraction Mode}
\label{sec:type-centric-mode}
For medium- and high-complexity inputs, \textsc{SMADE-IE} instantiates Type-Specific Agents for candidate types $\mathcal{A}$ selected by the Adaptive Mode Selector in Eq.~(\ref{eq:selector}) and adds a Review Agent to recover residual-type candidates. To resolve conflicts among agents, we introduce the Evidence-Driven Debate module that limits unconstrained discussion and weak-candidate interference, framing conflict resolution as evidence-based confidence comparison.

\subsubsection{Type-Specific Agent}
For each candidate type $t_\mathcal{A}^i \in \mathcal{A}$, we design a dedicated Type-Specific Agent with prompt $P_\mathcal{A}^i$. Given the input sentence $s$, the agent extracts candidate entities $e^i_\mathcal{A}$ associated with type $t_\mathcal{A}^i$:
\begin{eqnarray}\label{eq:type-agent}
e^i_\mathcal{A} &=& \mathrm{LLM}(s, P_\mathcal{A}^i),
\end{eqnarray}
where each element $e_\mathcal{A}^{i,j} \in e_\mathcal{A}^i$ denotes a candidate entity extracted for type $t_\mathcal{A}^i$. As illustrated in Figure~\ref{fig:overview}, the \textsc{Person} agent extracts both \textbf{Amazon} and \textbf{Jordan Lee}. We then assign each extracted entity $e_\mathcal{A}^{i,j}$ to its corresponding type $t_\mathcal{A}^i$ and aggregate all predictions to obtain the final output of the Type-Specific Agents, denoted as $E_\mathcal{A}$.

\subsubsection{Review Agent}
The Adaptive Mode Selector estimates $\mathcal{A}$ in a single recall-oriented pass, but may miss peripheral types in long or context-rich sentences. To address this issue, we introduce the Review Agent over the residual type set $\mathcal{A}^* = \mathcal{T}_e \setminus \mathcal{A}$. Let $P_{\mathcal{A}^*}^{\mathrm{rev}}$ denote the Review Agent prompt restricted to $\mathcal{A}^*$. The Review Agent produces a residual candidate set:
\begin{equation}\label{eq:review}
E_\mathcal{A}^* = \mathrm{LLM}(s, P_{\mathcal{A}^*}^{\mathrm{rev}}),
\end{equation}
The final output is defined as the union of the initial and residual sets:
\begin{equation}
\tilde{E}_\mathcal{A} = E_\mathcal{A} \cup E_\mathcal{A}^*.
\end{equation}
where each element in $\tilde{E}_\mathcal{A}$ is represented as a tuple $(\tilde{e}^i, \tilde{t}^i)$ containing a candidate entity and its assigned type. Since different Type-Specific Agents may assign different types to the same entity, we perform conflict detection before downstream processing. Specifically, if an entity $\tilde{e}^i$ is assigned multiple types, it is placed into the conflict set $\tilde{E}_{\mathcal{A}}^{\mathrm{conf}}$; otherwise, it is placed into the clean set $\tilde{E}_{\mathcal{A}}^{\mathrm{clean}}$:

\begin{equation}
\small
\begin{aligned}
\mathcal{Y}_{\tilde e}
&=
\{\tilde t \mid (\tilde e,\tilde t)\in \tilde E_{\mathcal A}\},\\
\tilde{E}^{\mathrm{clean}}_{\mathcal{A}}
&=
\{ (\tilde{e}, \tilde{t}) 
\mid |\mathcal{Y}_{\tilde e}| = 1 \},\\
\tilde{E}^{\mathrm{conf}}_{\mathcal{A}}
&=
\{ (\tilde e, \mathcal{Y}_{\tilde e})
\mid |\mathcal{Y}_{\tilde e}| > 1 \}.
\end{aligned}
\end{equation}

The clean set $\tilde{E}^{\mathrm{clean}}_{\mathcal{A}}$ is directly output, while the conflict set $\tilde{E}_{\mathcal{A}}^{\mathrm{conf}}$ is forwarded to the Evidence-Driven Debate module for conflict resolution and final type confirmation. Each element in $\tilde{E}_{\mathcal{A}}^{\mathrm{conf}}$ is represented as $(\tilde{e}^i, \{\tilde{t}^i_1, \dots, \tilde{t}^i_N\})$, where $\tilde{e}^i$ denotes a candidate entity and $\{\tilde{t}^i_1, \dots, \tilde{t}^i_N\}$ are its conflicting type assignments, and $N$ is the number of conflicting types assigned to the entity.

\subsubsection{Evidence-Driven Debate}
\label{sec:debate}

This module enhances the robustness and evidence-groundedness of conflict adjudication by mitigating unsupported reasoning and low-confidence interference in free-form multi-agent debates. Specifically, each conflicting type assignment is recast as a Toulmin-style argument to enable structured evidence inspection. An external evidence scorer estimates candidate confidence from sentence-level support, ranks candidates for pre-debate filtering, and computes evidence-based attack scores during the debate. Support and refutation signals are then aggregated through Beta updates, and the debate terminates based on two criteria: posterior stability and posterior superiority.

\paragraph{Toulmin-Guided Argument Construction}

\begin{figure}[t]
    \centering
    \includegraphics[scale=.40]{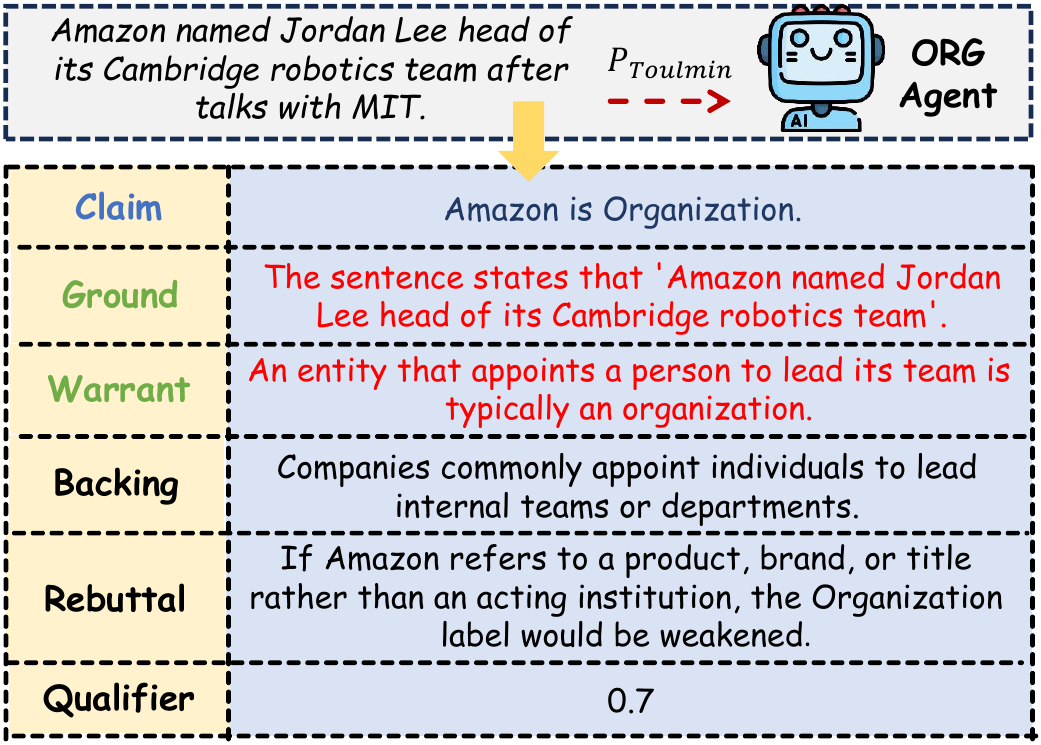}
    \caption{Example of a Toulmin-guided argument for the contested entity ``Amazon'', generated by the \textsc{Organization} agent.}
    \label{fig:toulmin_example}
    \vspace{-4mm}
\end{figure}



Recent work uses Toulmin theory to make implicit arguments explicit and improve the logical validity of LLM reasoning~\citep{gupta-etal-2024-harnessing,xiao-etal-2024-prove}. Following this paradigm, we construct a structured argument for each conflicting candidate entity $\tilde{e}^i \in \tilde{E}_\mathcal{A}^{\mathrm{conf}}$ and its associated types $\{\tilde{t}^i_1, \dots, \tilde{t}^i_N\}$ to constrain LLM discourse. Each agent organizes its argument into five components: Claim ($C$), Ground ($G$), Warrant ($W$), Backing ($B$), and Rebuttal ($R$). For a conflicting entity $\tilde{e}^i$, the Toulmin-guided argument for conflict type $\tilde{t}^i_n$ is defined as:
\begin{equation}\label{eq:toulmin}
\small
(C_n, G_n, W_n, B_n, R_n) = 
\mathrm{LLM}(s, \tilde{e}^i, \tilde{t}^i_n, P_{\mathrm{Toulmin}}),
\end{equation}
where $C_n$ is the claim for $\tilde{t}^i_n$, $G_n$ provides evidence from sentence $s$, $W_n$ explains how $G_n$ supports $C_n$, $B_n$ gives auxiliary support, and $R_n$ states a possible rebuttal. To measure how well $s$ supports each argument, we compute a confidence score $Q_n$ using an evidence scorer $\phi$~\citep{zha-etal-2023-alignscore}. Specifically, we concatenate $\{C_n, G_n, W_n, B_n\}$ into $\mathrm{Para}_n$ and construct a shared context $\hat{s}$ from sentence $s$ and all conflicting type definitions  $\{\tilde{t}^i_1, \dots, \tilde{t}^i_N\}$,
\begin{equation}
Q_n = \phi(\hat{s}, \mathrm{Para}_n) \in [0,1],
\end{equation}
where $\phi$ denotes an external evidence scorer~\citep{zha-etal-2023-alignscore} that measures how strongly the textual evidence $(G_n, W_n, B_n)$ supports the claim $C_n$. Each agent is represented as the tuple $\mathcal{D}_n = \{C_n, G_n, W_n, B_n, R_n, Q_n\}$, which forms a structured, evidence-grounded argument. As shown in Figure~\ref{fig:toulmin_example}, an \textsc{Organization} agent may use this format to justify the contested entity \textbf{Amazon}. Considering all candidate types for a contested entity $\tilde{e}^i$ incurs $\mathcal{O}(N^2)$ pairwise attacks per debate round. To reduce computational complexity and mitigate noise from weak candidates, we rank the $N$ Toulmin-guided arguments $\{\mathcal{D}_n\}_{n=1}^{N}$ by their qualifiers $Q_n$ and retain only the top two:
\begin{equation}
\small
\mathcal{D}_L
=
(\mathcal{D}_{\mathrm{top1}}, \mathcal{D}_{\mathrm{top2}})
=
\operatorname{Top2}_{\mathcal{D}_n}
\left(
\{\mathcal{D}_n\}_{n=1}^{N};
\ Q_n
\right),
\end{equation}
where $L=\{\mathrm{top1},\mathrm{top2}\}$ indexes the two retained candidates.
\paragraph{Bayesian Confidence Modeling}
We model candidate confidence with a Beta posterior
$\mathrm{Beta}(\alpha_l, \beta_l)$ for each retained candidate
$\mathcal{D}_l$, where $l \in L$ indexes the two candidates retained
after the Top-2 selection. The posterior accumulates supporting
($\alpha_l$) and refuting ($\beta_l$) evidence across debate iterations,
producing calibrated scores comparable across candidates. For candidate
$l$, we compute an alignment-based scaling factor:
\begin{align}
\rho_l &= 1 - \phi(x_l, R_l), \nonumber   \\
\kappa_l &= \kappa_{\min} + \rho_l (\kappa_{\max} - \kappa_{\min}),
\label{eq:kappa} 
\end{align}
where $x_l$ concatenates $s$ with the candidate type definition $\tilde{t}^i_l$, and $\kappa_l$ controls prior strength. Higher $\phi(x_l, R_l)$ implies weaker self-criticism and a smaller $\kappa_l$. The posterior is initialized with $Q_l$:
\begin{equation}
\alpha_l^{(0)} = Q_l \kappa_l, \qquad
\beta_l^{(0)} = (1-Q_l) \kappa_l,
\end{equation}
The initialized posterior parameters are then updated through the joint evidence and confidence updating process described below.

\paragraph{Evidence-Grounded Adversarial Refutation}
To ensure evidence-grounded refutations, we restrict the attack surface to the two Toulmin components that directly support candidate $l$: Ground ($G_l$) and Warrant ($W_l$). Ground provides supporting evidence, while Warrant links the evidence to the claim $C_l$; other components offer contextual or auxiliary justification. Let $m \in \{G, W\}$ denote an attackable component, and let $z_{l,m}$ be its textual content for candidate $l$. Each component has a validity score $v_{l,m}^{(t)} \in [0,1]$, initialized as $v_{l,m}^{(0)} = 1$, indicating its remaining credibility at iteration $t$. A component remains attackable while $v_{l,m}^{(t)} > 0$ and is removed once its validity reaches zero:
\begin{equation}
\label{eq:active-set}
\mathcal{V}_{l}^{(t)}
=
\left\{
m \in \{G, W\}
\;\middle|\;
v_{l,m}^{(t)} > 0
\right\}.
\end{equation}

At iteration $t$, the opposing agent $\mathcal{D}_{\bar{l}}$ attacks the defended agent $\mathcal{D}_{l}$ using sentence $s$, the active component set $\mathcal{V}_{l}^{(t)}$, and the directed attack prompt $P_{\bar{l}}^{\mathrm{atk}}$. For each defended component $m \in \mathcal{V}_{l}^{(t)}$, it generates a refutation $u_{\bar{l},m}$:
\begin{equation}
u_{\bar{l},m}
=
\mathrm{LLM}
\left(
s,\,
\mathcal{V}_{l}^{(t)},\,
P_{\bar{l},m}^{\mathrm{atk}}
\right).
\end{equation}
We measure attack effectiveness by how much the refutation weakens the defended component $z_{l,m}$ under the type-specific context $x_l$. The attack strength at iteration $t$ is
\begin{equation}
a_{l,m}^{(t)}
=
\sigma
\left(
\gamma
\left[
\phi(x_l,\, u_{\bar{l},m})
-
\phi(x_l,\, z_{l,m})
\right]
\right),
\end{equation}
where $\gamma$ is a temperature parameter and $\sigma(\cdot)$ is the sigmoid function. Larger $a_{l,m}^{(t)}$ indicates stronger support for the refutation relative to the defended content.

\begin{table*}[t]
\centering
\small
\setlength{\tabcolsep}{4.5pt}
\begin{tabular}{lcccccccccccc}
\toprule
\textbf{Method} & \multicolumn{2}{c}{\textbf{CoNLL03} (3)} & \multicolumn{2}{c}{\textbf{OntoNotes5 (18)}} & \multicolumn{2}{c}{\textbf{SciERC} (5)} & \multicolumn{2}{c}{\textbf{CrossRE} (38)} & \multicolumn{2}{c}{\textbf{REDFM} (10)} & \multicolumn{2}{c}{\textbf{Avg.}} \\
\cmidrule(lr){2-3} \cmidrule(lr){4-5} \cmidrule(lr){6-7} \cmidrule(lr){8-9} \cmidrule(lr){10-11} \cmidrule(lr){12-13}
 & $\text{F1}_\text{P}$ & $\text{F1}_\text{S}$ & $\text{F1}_\text{P}$ & $\text{F1}_\text{S}$ & $\text{F1}_\text{P}$ & $\text{F1}_\text{S}$ & $\text{F1}_\text{P}$ & $\text{F1}_\text{S}$ & $\text{F1}_\text{P}$ & $\text{F1}_\text{S}$ & $\text{F1}_\text{P}$ & $\text{F1}_\text{S}$ \\
\midrule
AEiO $^\diamondsuit$             & 65.24 & 61.17 & \underline{56.88} & \underline{43.12} & 35.36 & 23.80 & 34.61 & 30.80 & 21.14 & 17.45 & 42.65 & 35.27 \\
One-Step $^\clubsuit$          & \underline{77.29} & \underline{73.02} & 32.38 & 25.72 & 27.89 & 21.77 & 25.45 & 23.65 & 25.21 & 18.93 & 37.64 & 32.62 \\
G\&O $^\clubsuit$             & 76.36 & 72.35 & 31.65 & 23.40 & 31.83 & 24.05 & 27.00 & 25.38 & 27.07 & 20.61 & 38.78 & 33.16 \\
\textsc{CrossAgentIE}$^\spadesuit$ & 73.15 & 70.93 & 45.30 & 37.18 & \underline{35.96} & \underline{25.36} & \underline{46.15} & \underline{44.40} & \underline{29.20} & \underline{25.16} & \underline{45.95} & \underline{40.61} \\
\textsc{SMADE-IE}$^\spadesuit$(Ours)  & \textbf{77.33} & \textbf{74.72} & \textbf{66.08} & \textbf{52.05} & \textbf{46.56} & \textbf{37.80} & \textbf{65.50} & \textbf{60.56} & \textbf{31.12} & \textbf{27.13} & \textbf{57.32} & \textbf{50.45} \\
\bottomrule
\end{tabular}
\caption{
NER results on five benchmark datasets. Parentheses indicate the number of entity types (e.g., CoNLL03 has 3 entity types). \textbf{Bold} and \underline{underlined} values denote the best and second-best results, respectively. $\diamondsuit$, $\clubsuit$, and $\spadesuit$ indicate monolithic prompting, each-type prompting, and multi-agent debate methods, respectively.
}
\label{tab:ner-main}
\end{table*}

\begin{table*}[t]
\centering
\small
\setlength{\tabcolsep}{4.5pt}
\begin{tabular}{lcccccccccccc}
\toprule
\textbf{Method} & \multicolumn{2}{c}{\textbf{DocRED (32)}} & \multicolumn{2}{c}{\textbf{SemEval2010} (9)} & \multicolumn{2}{c}{\textbf{SciERC} (8)} & \multicolumn{2}{c}{\textbf{CrossRE} (17)} & \multicolumn{2}{c}{\textbf{REDFM} (32)} & \multicolumn{2}{c}{\textbf{Avg.}} \\
\cmidrule(lr){2-3} \cmidrule(lr){4-5} \cmidrule(lr){6-7} \cmidrule(lr){8-9} \cmidrule(lr){10-11} \cmidrule(lr){12-13}
 & $\text{F1}_\text{P}$ & $\text{F1}_\text{S}$ & $\text{F1}_\text{P}$ & $\text{F1}_\text{S}$ & $\text{F1}_\text{P}$ & $\text{F1}_\text{S}$ & $\text{F1}_\text{P}$ & $\text{F1}_\text{S}$ & $\text{F1}_\text{P}$ & $\text{F1}_\text{S}$ & $\text{F1}_\text{P}$ & $\text{F1}_\text{S}$ \\
\midrule
AEiO $^\diamondsuit$             & \underline{30.32} & \underline{24.84} & \underline{27.55} & \underline{9.86} & \underline{19.98} & \underline{10.91} & 4.79 & 3.56 & \underline{10.04} & \underline{6.24} & \underline{18.54} & \underline{11.08} \\
One-Step $^\clubsuit$         & 22.15 & 18.41 & 16.13 & 5.16 & 18.56 & 8.94 & \underline{12.00} & 8.31 & 7.11 & 5.34 & 15.19 & 9.23 \\
G\&O $^\clubsuit$            & 19.69 & 16.71 & 12.16 & 4.56 & 13.07 & 5.58 & 8.90 & 7.01 & 5.65 & 4.12 & 11.89 & 7.60 \\
\textsc{CrossAgentIE} $^\spadesuit$ & 13.72 & 11.49 & 11.77 & 2.91 & 14.61 & 8.79 & 11.81 & \underline{8.57} & 5.36 & 3.44 & 11.45 & 7.04 \\
\textsc{SMADE-IE} $^\spadesuit$  & \textbf{30.78} & \textbf{26.99} & \textbf{29.54} & \textbf{16.76} & \textbf{21.47} & \textbf{11.20} & \textbf{16.22} & \textbf{12.25} & \textbf{13.83} & \textbf{11.34} & \textbf{22.37} & \textbf{15.71} \\
\bottomrule
\end{tabular}
\caption{RE results on five benchmark datasets. \textsc{SMADE-IE} obtains the strongest macro average across all RE benchmarks. 
}
\label{tab:re-main}
\end{table*}

\begin{table}[t]
\centering
\small
\setlength{\tabcolsep}{1.5pt}
\begin{tabular}{lcccccc}
\toprule
\textbf{Method} & \multicolumn{2}{c}{\textbf{CoNLL04} (5)} & \multicolumn{2}{c}{\textbf{NYT} (7)} & \multicolumn{2}{c}{\textbf{Avg.}} \\
\cmidrule(lr){2-3} \cmidrule(lr){4-5} \cmidrule(lr){6-7}
 & $\text{F1}_\text{P}$ & $\text{F1}_\text{S}$ & $\text{F1}_\text{P}$ & $\text{F1}_\text{S}$ & $\text{F1}_\text{P}$ & $\text{F1}_\text{S}$ \\
\midrule
\textsc{CrossAgentIE} $^\spadesuit$ & 42.50 & 29.05 & 17.23 & 16.77 & 29.87 & 22.91 \\
\textsc{SMADE-IE} $^\spadesuit$        & \textbf{58.44} & \textbf{44.74} & \textbf{30.22} & \textbf{29.22} &\textbf{44.33} &\textbf{36.98}\\
\bottomrule
\end{tabular}
\caption{JERE results on two benchmark datasets.
The head and tail entity types and the relation type must all match.}
\vspace{-2mm}
\label{tab:jere-main}
\end{table}

\paragraph{Joint Evidence and Confidence Updating}
To prevent posterior recovery through reformulated invalid components, we couple component validity with the Beta posterior via the shared attack score $a_{l,m}^{(t)}$. This ties posterior updates to evidence credibility and keeps both update processes consistent. For each component, $v_{l,m}^{(t)}$ is updated in two stages. First, the attack induces exponential decay:

\begin{equation}
\tilde{v}_{l,m}^{(t+1)}
=
v_{l,m}^{(t)}
\exp\!\left(-\eta\,a_{l,m}^{(t)}\right),
\end{equation}
where $\eta > 0$ is the decay rate. Given a validity activity threshold $\theta_v$, the realized validity is
\begin{equation}
\small
\label{eq:vupdate}
v_{l,m}^{(t+1)} =
\begin{cases}
\tilde{v}_{l,m}^{(t+1)},
& \tilde{v}_{l,m}^{(t+1)} > \theta_v, \\[6pt]
\dfrac{\theta_v + 1}{2}\,\omega^{k^l_{m}},
&
\begin{aligned}
0 < \tilde{v}_{l,m}^{(t+1)}
&\leq \theta_v, \\
\text{and } 
\dfrac{\theta_v + 1}{2}\,\omega^{k^l_{m}}
&\geq \theta_v,
\end{aligned}
\\[10pt]
0,
& \text{otherwise}.
\end{cases}
\end{equation}
where $\tilde{v}_{l,m}^{(t+1)}$ is retained if it remains above $\theta_v$. Otherwise, the component enters revision and is reset to the discounted ceiling $\frac{\theta_v+1}{2}\omega^{k^l_m}$, where $k^l_m$ is the number of prior revisions and $\omega \in (0,1)$ is the revision discount factor. Repeated revisions therefore reduce credibility. If the revised validity still exceeds $\theta_v$, the component remains in $\mathcal{V}_l$; otherwise, it is permanently removed from the attack surface. The same attack score updates the Beta posterior of candidate $l$ by adding refutation mass to $\beta_l$ and residual support mass to $\alpha_l$:

{\small
\begin{equation}
\label{eq:beta-update}
\begin{aligned}
\beta_l^{(t+1)}
&=
\beta_l^{(t)}
+
\frac{\bar{v}_{\bar{l}}^{(t)}}{\left|\mathcal{V}_l^{(t)}\right|}
\sum_{m\in\mathcal{V}_l^{(t)}}
{a}_{l,m}^{(t)} \cdot{v}_{l,m}^{(t)} \\
\alpha_l^{(t+1)}
&=
\alpha_l^{(t)}
+
\frac{1}{\left|\mathcal{V}_l^{(t)}\right|}
\sum_{m\in\mathcal{V}_l^{(t)}}
\big(1-a_{l,m}^{(t)}\big) \cdot v_{l,m}^{(t)}
\end{aligned}
\end{equation}
}
where $\bar{v}_{\bar{l}}^{(t)}$ denotes the average validity of the opponent's active evidence components:
\begin{equation}
\bar{v}_{\bar{l}}^{(t)}
=
\frac{1}{|\mathcal{V}_{\bar{l}}^{(t)}|}
\sum_{m \in \mathcal{V}_{\bar{l}}^{(t)}}
v_{\bar{l},m}^{(t)} .
\end{equation}
The weights have distinct roles: $\bar{v}_{\bar{l}}^{(t)}$ downweights attacks from opponents with weakened evidence, while $v_{l,m}^{(t)}$ reduces the influence of components weakened in earlier rounds.

\paragraph{Dual-Track Early Stopping}


To avoid redundant debate rounds, we terminate the process when either \textbf{posterior stability} or \textbf{posterior superiority} is satisfied. Both criteria operate on the Beta posteriors induced by the joint update rule:
\begin{equation}
\pi_l^{(t)} \sim \mathrm{Beta}\!\left(\alpha_l^{(t)}, \beta_l^{(t)}\right),
\end{equation}
for each candidate $l\in L$ at iteration $t$. The \textbf{posterior stability} criterion measures convergence between consecutive posterior distributions using the average squared Hellinger distance:
\begin{equation}\label{eq:stab}
\begin{aligned}
\Delta_{\mathrm{conf}}^{(t)} =
\frac{1}{2}\sum_{l\in L}
H^2\!\Big( \pi_l^{(t)},\pi_l^{(t-1)} \Big)
< \varepsilon,
\end{aligned}
\end{equation}
where $H^2(\cdot)$ denotes squared Hellinger distance. The debate terminates once $\Delta_{\mathrm{conf}}^{(t)} < \varepsilon$, indicating that the posteriors have effectively converged.

The \textbf{posterior superiority} criterion stops the debate when the leading candidate is sufficiently likely to outperform its opponent. For scalar comparison, we use the posterior mean
\begin{equation}
\label{eq:posterior-mean}
\hat{p}_l^{(t)} =
\frac{\alpha_l^{(t)}}{\alpha_l^{(t)} + \beta_l^{(t)}},
\end{equation}
which integrates all supporting and refuting evidence accumulated up to iteration $t$. Let
\begin{equation}
l^\ast=\arg\max_{l\in L}\hat{p}_l^{(t)},
\end{equation}
where $l^\ast$ denotes the leading candidate at iteration $t$, and $\bar{l}^\ast$ denotes its opponent. Early termination is triggered once
\begin{equation}\label{eq:psup}
\mathcal{P}\!\left(\pi_{l^\ast}^{(t)}>\pi_{\bar{l}^\ast}^{(t)}\right)
>
\delta_{\mathrm{stop}},
\end{equation}
indicating sufficiently strong evidence in favor of the leading candidate.
When neither stopping criterion is met before reaching the maximum round budget $T_{\max}$, the final winner is determined by the external evidence scorer $\phi$:
\begin{equation}
l^{\mathrm{winner}}
=
\arg\max_{l\in L}
\phi(\hat{s},\mathrm{Para}_l).
\end{equation}
Appendix~\ref{app:bound} provides a Chebyshev-style bound motivating the dual-track stopping rule.
\subsection{Iterative Entity--Relation Alignment}
\label{sec:alignment}
\begin{figure}[t]
  \centering
  \includegraphics[scale=.65]{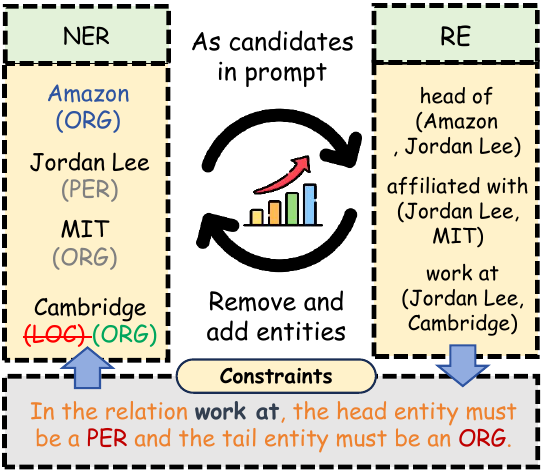}
  \caption{Iterative Entity--Relation Alignment under ontology constraints.}
  \label{fig:IERA}
  \vspace{-6mm}
\end{figure}




For JERE, we enforce ontology-consistent alignment between $E$ and $R$, as shown in Figure~\ref{fig:IERA}. At iteration $t$, the procedure proceeds as follows.

\paragraph{Entity completion}
Given the candidate relation set $\bar{R}^{(t)} =
RE(E^{(t)})$, we add entities that appear in $\bar{R}^{(t)}$ but are missing from $E^{(t)}$ after inferring their types:
\begin{equation}
\small
\bar{E}^{(t+1)} =
E^{(t)} \cup
\{ (e,t)\notin E^{(t)} \mid e \in \xi(\bar{R}^{(t)})\},
\end{equation}
where $\xi(\cdot)$ denotes the entities involved in the candidate relations. This step completes missing entities under relational evidence.

\paragraph{Consistency correction}
We further enforce ontology consistency on $\bar{R}^{(t)}$ with respect to $\bar{E}^{(t+1)}$.
For each violation, the Consistency Agent either revises the conflicting entity type if the relation type is trusted, or blacklists the relation otherwise.
Let $C_-^{(t)}$ denote the conflicting entity--type assignments to be replaced and $C_+^{(t)}$ denote their corrected assignments returned by the Consistency Agent. The entity set is first corrected as:
\begin{equation}
\bar{E}^{(t+1)} =
\left(\bar{E}^{(t+1)} \setminus C_-^{(t)}\right)
\cup C_+^{(t)} .
\end{equation}
The aligned relations and entities are updated as:
\begin{equation}
R^{(t)} = \bar{R}^{(t)} \setminus B^{(t)} .
\end{equation}
Entities not involved in any remaining relation are removed:
\begin{equation}
E^{(t+1)} =
\{ (e,t) \in \bar{E}^{(t+1)} \mid e \in \xi(R^{(t)}) \}.
\end{equation}
We regenerate candidate relations from the updated entity set and filter them using the updated blacklist:
\begin{equation}
R^{(t+1)} =
\bar{R}^{(t+1)} \setminus B^{(t+1)} .
\end{equation}
The procedure terminates once all violations are resolved or $K_{\mathrm{JERE}}$ is reached, yielding the aligned JERE output.

\section{Experiment and Analysis}
This section presents the main experimental results and analyses. We report micro Partial F1 ($\mathrm{F1}_\mathrm{P}$) and Strict F1 ($\mathrm{F1}_\mathrm{S}$), together with averages across datasets. Partial F1 allows boundary overlap between predicted and gold spans, whereas Strict F1 requires exact span matching.  Additional details on datasets, evaluation metrics, baselines, implementation settings, hyperparameters, and prompts are provided in Appendices~\ref{app:experiment}--\ref{app:prompts}.

\subsection{Main Results}

Tables~\ref{tab:ner-main} and~\ref{tab:re-main} report NER and RE results on five benchmarks using GPT-3.5-Turbo-0125 as the backbone. \textsc{SMADE-IE} achieves the best performance across both tasks, demonstrating the effectiveness of multi-agent deliberation for information extraction. For NER, \textsc{SMADE-IE} outperforms all baselines, with average $\mathrm{F1}_\mathrm{P}$ gains of 14.67 over AEiO, 19.68 over One-Step, 18.54 over G\&O, and 11.37 over \textsc{CrossAgentIE}. The gains are larger on datasets with richer type inventories, surpassing \textsc{CrossAgentIE} by 20.78 $\mathrm{F1}_\mathrm{P}$ on OntoNotes5, 19.35 on CrossRE, and 10.60 on SciERC. For RE, \textsc{SMADE-IE} obtains the highest $\mathrm{F1}_\mathrm{P}$ and $\mathrm{F1}_\mathrm{S}$ scores on all benchmarks, improving average $\mathrm{F1}_\mathrm{P}$ by 3.83, 7.18, 10.48, and 10.92 over AEiO, One-Step, G\&O, and \textsc{CrossAgentIE}, respectively. Its largest gain over \textsc{CrossAgentIE} occurs on SemEval2010 (+17.77 $\mathrm{F1}_\mathrm{P}$), while consistent improvements on DocRED, SciERC, CrossRE, and REDFM indicate strong generalizability across relation extraction settings.

For the JERE task, which requires joint entity-relation extraction and cross-task completion, we primarily compare with \textsc{CrossAgentIE}. As shown in Table~\ref{tab:jere-main}, \textsc{SMADE-IE} improves over \textsc{CrossAgentIE} by 14.46 $\mathrm{F1}_\mathrm{P}$ and 14.07 $\mathrm{F1}_\mathrm{S}$ on average. The gains are consistent on both datasets, with $\mathrm{F1}_\mathrm{P}$ increasing from 42.50 to 58.44 on CoNLL04 and from 17.23 to 30.22 on NYT.

\subsection{Ablation Studies}
\label{sec:ablation}

\begin{table*}[t]
\centering
\small
\setlength{\tabcolsep}{9.0pt}
\begin{tabular}{lcccccccc}
\toprule
\multirow{2}{*}{\textbf{Variant}}
& \multicolumn{2}{c}{\textbf{SciERC (NER)}}
& \multicolumn{2}{c}{\textbf{CrossRE (RE)}}
& \multicolumn{2}{c}{\textbf{CoNLL04 (JERE)}}
& \multicolumn{2}{c}{\textbf{NYT (JERE)}} \\
\cmidrule(lr){2-3}
\cmidrule(lr){4-5}
\cmidrule(lr){6-7}
\cmidrule(lr){8-9}
& $\mathrm{F1}_{\mathrm{P}}$ & $\mathrm{F1}_{\mathrm{S}}$
& $\mathrm{F1}_{\mathrm{P}}$ & $\mathrm{F1}_{\mathrm{S}}$
& $\mathrm{F1}_{\mathrm{P}}$ & $\mathrm{F1}_{\mathrm{S}}$
& $\mathrm{F1}_{\mathrm{P}}$ & $\mathrm{F1}_{\mathrm{S}}$ \\
\midrule

\textbf{Full \textsc{SMADE-IE}}
& 46.56 & \textbf{37.80}
& \textbf{16.22} & \textbf{12.25}
& 58.44 & \textbf{44.74}
& \textbf{30.22} & \textbf{29.22} \\

~~~~w/o \emph{IERA}
& -- & --
& -- & --
& 59.03 & 42.73
& 24.94 & 22.47 \\

\midrule
\textbf{Type-Centric Only}
& \textbf{46.75} & 36.81
& 15.73 & 11.11
& \textbf{59.43} & 40.33
& 29.65 & 28.47 \\

~~~~w/o \emph{Relevant Type Selection}
& 44.16 & 33.71
& 14.00 & 10.81
& 48.66 & 34.18
& 27.43 & 26.43 \\

~~~~w/o \emph{Review Agent}
& 45.70 & 35.03
& 14.10 & 9.13
& 52.75 & 35.12
& 29.50 & 27.56 \\

~~~~w/o \emph{Evidence-Driven Debate}
& 45.86 & 34.98
& 13.22 & 10.09
& 54.13 & 38.02
& 24.24 & 23.38 \\

\midrule
\textbf{Global Only}
& 46.15 & 33.87
& 14.24 & 10.09
& 52.47 & 40.20
& 16.28 & 14.27 \\

~~~~w/o \emph{Verification Agent}
& 43.37 & 33.03
& 12.83 & 9.06
& 46.92 & 35.19
& 12.02 & 11.16 \\

\bottomrule
\end{tabular}
\caption{
Ablation results across NER, RE, and JERE. The first block evaluates IERA, while the remaining blocks analyze Relevant Type Selection, Review Agent, Evidence-Driven Debate, and Verification Agent. ``--'' denotes inapplicable settings.
}
\label{tab:ablation}
\vspace{-3mm}
\end{table*}
Table~\ref{tab:ablation} evaluates the contribution of individual components across NER, RE, and JERE. On SciERC NER, the full \textsc{SMADE-IE} achieves an $\mathrm{F1}_{\mathrm{P}}$ of 46.56 and the highest $\mathrm{F1}_{\mathrm{S}}$ of 37.80. Compared with \emph{Type-Centric Only}, removing \emph{Relevant Type Selection}, \emph{Review Agent}, and \emph{Evidence-Driven Debate} consistently reduces $\mathrm{F1}_{\mathrm{S}}$ (36.81$\rightarrow$33.71, 35.03, and 34.98, respectively). In \emph{Global Extraction Mode}, removing the \emph{Verification Agent} also degrades performance ($\mathrm{F1}_{\mathrm{P}}$: 46.15$\rightarrow$43.37; $\mathrm{F1}_{\mathrm{S}}$: 33.87$\rightarrow$33.03), further supporting the effectiveness of verification for lightweight extraction.

A similar trend is observed on CrossRE RE. The full model achieves the best performance with 16.22 $\mathrm{F1}_{\mathrm{P}}$ and 12.25 $\mathrm{F1}_{\mathrm{S}}$. Compared with \emph{Type-Centric Only}, removing \emph{Relevant Type Selection}, \emph{Review Agent}, and \emph{Evidence-Driven Debate} reduces $\mathrm{F1}_{\mathrm{S}}$ (11.11$\rightarrow$10.81, 9.13, and 10.09, respectively). Removing the \emph{Verification Agent} from \emph{Global Extraction Mode} likewise decreases performance ($\mathrm{F1}_{\mathrm{P}}$: 14.24$\rightarrow$12.83; $\mathrm{F1}_{\mathrm{S}}$: 10.09$\rightarrow$9.06). These results demonstrate that the proposed components remain effective for standalone NER and RE tasks rather than being specific to joint extraction.

For JERE, removing \emph{IERA} substantially degrades performance on NYT ($\mathrm{F1}_{\mathrm{P}}$: 30.22$\rightarrow$24.94; $\mathrm{F1}_{\mathrm{S}}$: 29.22$\rightarrow$22.47), with a smaller effect on CoNLL04, indicating that IERA is beneficial when entity--relation inconsistencies are frequent. Comparisons with the single-mode variants further validate the effectiveness of \emph{Adaptive Mode Selector}: compared with \emph{Global Only}, the full model improves $\mathrm{F1}_{\mathrm{S}}$ (40.20$\rightarrow$44.74 on CoNLL04; 14.27$\rightarrow$29.22 on NYT), while compared with \emph{Type-Centric Only}, it yields consistent gains (40.33$\rightarrow$44.74 and 28.47$\rightarrow$29.22, respectively). Within each mode, the \emph{Verification Agent}, \emph{Relevant Type Selection}, \emph{Review Agent}, and \emph{Evidence-Driven Debate} all contribute, with Relevant Type Selection particularly important on CoNLL04 ($\mathrm{F1}_{\mathrm{P}}$: 59.43$\rightarrow$48.66 when removed) and Evidence-Driven Debate on NYT ($\mathrm{F1}_{\mathrm{P}}$: 29.65$\rightarrow$24.24; $\mathrm{F1}_{\mathrm{S}}$: 28.47$\rightarrow$23.38). These results confirm that each module plays a complementary role across extraction settings.

\subsection{Token Efficiency}

\begin{table}[t]
\centering
\setlength{\tabcolsep}{2.2pt}
\resizebox{\columnwidth}{!}{
\begin{tabular}{lcccc}
\toprule
\textbf{Method} 
& \multicolumn{2}{c}{\textbf{NER}} 
& \multicolumn{2}{c}{\textbf{RE}} \\
\cmidrule(lr){2-3} \cmidrule(lr){4-5}
& \textbf{CoNLL03} (3) & \textbf{REDFM} (10) 
& \textbf{DocRED} (32) & \textbf{CrossRE} (17) \\
\midrule
AEiO           & 1078 & 2163  & 5460  & 4798  \\
One-Step          & 1407 & 7673  & 7615  & 6846  \\
G\&O             & 2304 & 14202 & 9271  & 14312 \\
CrossAgentIE      & 1508 & 13480 & 21784 & 18923 \\
\textsc{SMADE-IE} & 1094 & 11514 & 3240  & 3765  \\
\bottomrule
\end{tabular}
}
\caption{Average total token cost per sample across different datasets.}
\label{tab:selected_token_cost}
\vspace{-6mm}
\end{table}

Table~\ref{tab:selected_token_cost} reports the average total token cost per sample on selected NER and RE datasets.
On CoNLL03, \textsc{SMADE-IE} uses 1,094 tokens, close to AEiO (1,078) and lower than One-Step (1,407), G\&O (2,304), and \textsc{CrossAgentIE} (1,508).
On REDFM NER, \textsc{SMADE-IE} also reduces cost compared with G\&O and \textsc{CrossAgentIE}, using 11,514 tokens versus 14,202 and 13,480.
The savings are more pronounced on RE: \textsc{SMADE-IE} uses 3,240 tokens on DocRED and 3,765 on CrossRE, far below \textsc{CrossAgentIE}'s 21,784 and 18,923 tokens.
It is also more efficient than One-Step and G\&O on both RE datasets.
These results show that sparse type selection avoids unnecessary full-schema agent calls, keeping the cost close to lightweight prompting on simple cases while substantially reducing multi-agent debate overhead on relation-heavy datasets.

\section{Conclusion}
This paper proposes \textsc{SMADE-IE}, a sparse and evidence-driven multi-agent framework for zero-shot information extraction. \textsc{SMADE-IE} employs the Adaptive Mode Selector to dynamically route inputs into the lightweight Global Extraction Mode or the Type-Centric Extraction Mode, reducing redundant agent interactions and reasoning noise. For conflicting predictions, we further introduce the Evidence-Driven Debate mechanism with Toulmin-style argument modeling and Bayesian confidence updates. In addition, the Iterative Entity--Relation Alignment strategy is designed for joint extraction tasks. Experimental results on 9 benchmark datasets across NER, RE, and JERE tasks demonstrate that \textsc{SMADE-IE} consistently outperforms existing zero-shot IE baselines while improving token efficiency. Future work will explore more efficient debate scheduling and broader structured extraction settings.

\section{Acknowledgements}
This research is supported by the National Natural Science Foundation of China (62476097, 62402184), Guangdong Provincial Fund for Basic and Applied Basic Research—Regional Joint Fund Project (Key Project) (2023B1515120078), Guangdong Provincial Natural
Science Foundation for Outstanding Youth Team Project (2024B1515040010), the Guangdong S\&T Programme (2025B0101120003), the Fundamental Research Funds for the Central Universities, South China University of Technology (x2rjD2250190). Li Yuan was funded by the School of Artificial Intelligence, South China Normal University.

\section*{Limitations}
\textsc{SMADE-IE} has two main limitations. First, Evidence-Driven Debate relies on a frozen external evidence scorer, whose errors may propagate to subsequent confidence aggregation. Second, due to the inference cost of multi-agent LLM evaluation, some datasets are evaluated on fixed sampled subsets. Although all methods use the same instances for fair comparison, this setting does not quantify sampling uncertainty.

\section*{Ethics Statement}
In this work, we propose \textsc{SMADE-IE}, a sparse multi-agent framework designed to improve large language model performance on fundamental information extraction tasks, including named entity recognition and relation extraction. All our analyses and evaluations are based on standard, publicly available academic datasets. We do not anticipate any specific ethical issues or negative societal impacts regarding the methodology or the topics of this research.

\bibliography{main}

\clearpage
\appendix

\section{Related Works}

\subsection{Supervision-based Information Extraction}
Early IE methods mainly rely on supervised learning, typically modeling NER as sequence labeling and RE as classification or structured prediction~\citep{lafferty2001conditional,mccallum2003early,kambhatla2004combining,rink2010utd}. Neural IE methods reduce reliance on manual feature engineering by learning distributed representations and contextual encoders for sequence labeling, relation classification, and joint extraction~\citep{zeng-etal-2014-relation,zheng-etal-2017-joint,yuan2024fine,yuan2024few}. More recently, pretrained-language-model fine-tuning and instruction-tuning methods have further improved IE performance under supervised and transfer settings~\citep{lu-etal-2022-unified,lou2023universalinformationextractionunified}.
However, these methods usually assume sufficient annotated data, stable domains, and fixed entity and relation schemas, making them difficult to adapt to annotation-scarce settings or dynamically changing schemas. Weakly and distantly supervised methods reduce annotation cost by using unlabeled corpora, external knowledge bases, or heuristic labeling functions~\citep{Li_2022,Chen_2023,gao-etal-2024-promptre},
but they still require additional supervision signals or manual design efforts and may suffer from noisy pseudo-labels~\citep{Lison_2021}. In contrast, this work focuses on zero-shot IE, where models perform extraction using only input text, type definitions, and output constraints. While zero-shot IE poses greater challenges than fully or semi-supervised methods, it offers superior adaptability to a dynamically changing world without requiring any additional fine-tuning.

\subsection{LLM-based Zero-shot Information Extraction}
Recent studies have explored the use of LLMs for zero-shot IE by prompting models with task instructions, type definitions, or output schemas~\citep{xu2024largelanguagemodelsgenerative,zhang-etal-2025-survey}. \emph{Monolithic-prompt} methods extract all elements in a single pass, offering high efficiency but often suffering from type confusion, boundary errors, and missed mentions as the schema size increases~\citep{xie-etal-2023-empirical,li2024gno,han2024empiricalstudyinformationextraction}. In contrast, \emph{each-type prompt} methods query entity or relation types separately, enabling more focused extraction but frequently producing overlapping or conflicting outputs across prompts~\citep{li-etal-2023-revisiting-large,li2024gno}.

Recent agentic LLM systems increasingly decompose complex tasks into specialized agents and leverage structured intermediate knowledge to improve reasoning and coordination~\citep{gong2026repomind,wang2025model}. In zero-shot IE, \emph{multi-agent debate} methods instantiate type-specific agents and resolve conflicting predictions through inter-agent deliberation. Related research has further explored multi-agent collaboration, tool use, knowledge augmentation, and workflow orchestration for IE~\citep{wang2024debateoptimizationadaptiveconformal,Shi_2024,wang2025cooperativemultiagentframeworkzeroshot,luo2025onekedockerizedschemaguidedllm}. For example, \textsc{CrossAgentIE}~\citep{crossagentie} assigns dedicated agents to entity and relation types and resolves structured prediction conflicts through cross-type and cross-task debate. However, these approaches select the full type schema for every input, causing token cost and contextual noise to scale with schema size. Moreover, their free-form debate mechanisms lack explicit modeling of disputed claims, evidence constraints, and confidence updates, which can lead to unreliable adjudication. In contrast, \textsc{SMADE-IE} selects only Router-selected types and replaces free-form deliberation with a Toulmin-structured, evidence-scored Bayesian debate framework.

\begin{table*}[t]
\centering
\small
\setlength{\tabcolsep}{8pt}
\begin{tabular}{lccccccc}
\toprule
\textbf{Statistic} & CoNLL03 & OntoNotes5 & CoNLL04 & NYT & CrossRE & REDFM & SciERC \\
\midrule
\# Selected Instances    & 300     & 200      & 288     & 369     & 200     & 446     & 269 \\
Avg.\ Sequence Length    & 13      & 18       & 29      & 36      & 37      & 76      & 23  \\
Avg.\ Entities per Instance  & 1.50  & 1.31  & 2.38  & 2.00  & 4.72  & 3.57  & 2.19 \\
Entity Types             & 3       & 18       & 3       & 3       & 38      & 10       & 5 \\
\bottomrule
\end{tabular}
\caption{Statistics of the NER benchmarks used in our evaluation. ``\# Selected Instances'' denotes the number of evaluation samples used in our experiments, ``Avg.\ Sequence Length'' indicates the average number of tokens per instance, ``Avg. Entities per Instance'' refers to the average number of entity mentions in each sample and ``Entity Types'' refers to the number of entity categories evaluated in each dataset.}
\label{tab:ner-stats}
\end{table*}

\begin{table*}[t]
\centering
\small
\setlength{\tabcolsep}{8pt}
\begin{tabular}{lccccccc}
\toprule
\textbf{Statistic} & DocRED & SemEval2010 & CoNLL04 & NYT & CrossRE & REDFM & SciERC \\
\midrule
\# Selected Instances & 200     & 200     & 288     & 369   & 200     & 446     & 269 \\
Avg.\ Sequence Length & 30      & 17      & 29      & 36    & 37      & 76      & 23  \\
Avg.\ Relations per Instance  & 2.46 & 0.84 & 1.41 & 0.72 & 5.05 & 2.77 & 1.88 \\
Relation Types & 32 & 9   & 5       & 7  & 17      & 32      & 8 \\
\bottomrule
\end{tabular}
\caption{Statistics of the RE benchmarks used in our evaluation. ``\# Selected Instances'' denotes the number of evaluation samples used in our experiments, ``Avg.\ Sequence Length'' indicates the average number of tokens per instance, ``Avg.\ Relations per Instance'' refers to the average number of relation mentions in each sample, and ``Relation Types'' refers to the number of relation categories evaluated in each dataset. All evaluated relations are binary.}
\label{tab:re-stats}
\end{table*}

\section{Experimental Setup}
\label{app:experiment}
\subsection{Datasets and Evaluation Metrics}

We evaluate \textsc{SMADE-IE} on 12 zero-shot IE task settings over 9 benchmark datasets, spanning NER, RE, and JERE across general-domain, scientific, and document-level text. For NER, we use CoNLL03~\citep{conll03}, OntoNotes5~\citep{ontonotes5}, SciERC~\citep{scierc}, CrossRE~\citep{crossre}, and REDFM~\citep{redfm}. For RE, we evaluate on DocRED~\citep{docred}, SemEval-2010 Task 8~\citep{semeval2010re}, and reuse SciERC, CrossRE, and REDFM. For JERE, we use CoNLL04~\citep{conll04} and NYT~\citep{nyt}. During inference, the model is provided only with natural-language ontology definitions as external knowledge, following the zero-shot evaluation protocol of \citet{li2024gno} and \citet{crossagentie}.

For datasets whose ontologies contain rare, ambiguous, or under-specified labels, we follow the evaluation settings adopted by prior work~\citep{li2024gno}. Specifically, CoNLL03 and CoNLL04 retain only the three concrete entity types. CrossRE retains 38 out of 39 entity types after removing the \textsc{Other} category and uses all 17 semantic relation types. REDFM and SciERC exclude only reserved generic labels. DocRED is restricted to the 32 most frequent relation types, while SemEval is limited to the 9 forward relation types. Detailed dataset entity-level statistics are shown in Table~\ref{tab:ner-stats} and relation-level statistics are shown in Table~\ref{tab:re-stats}. For datasets containing more than a few hundred test instances, we randomly sample a fixed evaluation subset shared across all baselines and ablation settings to ensure fair and directly comparable evaluations of token cost and F1 performance. The sampled instance IDs are released with our code for reproducibility.

\subsection{Baselines}
We compare \textsc{SMADE-IE} against four representative zero-shot LLM-based IE methods spanning three paradigms. For \emph{monolithic prompting}, \textbf{All-Entity-in-One (AEiO)}~\citep{li2024gno} extracts all candidate types in a single prompt. For \emph{each-type prompting}, \textbf{One-Step}~\citep{li2024gno} issues a dedicated prompt for each entity or relation type, while \textbf{Generation and Organization (G\&O)}~\citep{li2024gno} decouples content generation from output organization. For \emph{multi-agent debate}, \textbf{\textsc{CrossAgentIE}}~\citep{crossagentie} instantiates type-specific agents and resolves conflicts through cross-type and cross-task debate.

\subsection{Implementation Details}
\label{app:implementation}
Unless otherwise specified, all main experiments use
GPT-3.5-Turbo-0125 in a zero-shot setting, receiving only the input
text, task schema, and type definitions.
Agent interactions are implemented with AutoGen\footnote{\url{https://microsoft.github.io/autogen/}}, and the external evidence scorer $\phi$ uses the AlignScore-base checkpoint~\citep{zha-etal-2023-alignscore}\footnote{\url{https://huggingface.co/yzha/AlignScore}} deployed locally.
For \textsc{SMADE-IE}, the maximum number of debate rounds is three, with $\delta_{\mathrm{stop}}=0.75$, $\varepsilon=0.02$, and $\gamma=8$. All hyperparameters are fixed across the twelve evaluation settings; the complete configuration is reported in Table~\ref{tab:smade-hp}.

For baselines, we use the official implementations of AEiO, One-Step, and G\&O. We extend AEiO to RE by retaining its all-in-one chat structure and replacing only the output schema: a single GPT-3.5-Turbo-0125 session jointly extracts relation mentions, prunes ungrounded ones, and emits per-type Markdown tables with head, tail, and validity columns. Outputs are parsed by normalizing relation-type headers, discarding rows whose validity token is not ``yes'' or whose head and tail coincide, and mapping the rest to token spans via the spaCy and locate-phrase routines from the G\&O codebase.

\textsc{CrossAgentIE}~\citep{crossagentie} has no public code and is re-implemented on AutoGen with the same backbone. Each type is bound to a Type-Specific Agent that emits per-mention extractions with confidence scores via the original \texttt{\#\#\# head \#\#\#}/\texttt{@@@ tail @@@} markers. When a head--tail pair is claimed by more than one type, the top-two claimants by confidence enter a \texttt{GroupChat} debate of up to two rounds, settled once the leader-runner-up gap exceeds $0.1$; a Judge Agent breaks intra-group ties and a Consistency Agent reconciles cross-task conflicts. For joint extraction, the Consistency Agent runs one additional reassessment that drops or retypes flagged entities when its confidence falls below $0.75$. All agents use temperature $0.9$ and frequency penalty $0.1$.

\begin{table}[t]
\centering
\small
\setlength{\tabcolsep}{15pt}
\begin{tabular}{ll}
\toprule
\textbf{Symbol / Name} & \textbf{Value} \\
\midrule
$T_{\max}$ (max debate rounds) & $3$ \\
$[\kappa_{\min}, \kappa_{\max}]$ (prior strength) & $[0.8,\,1.5]$ \\
$\theta_v$ (validity activity threshold) & $0.6$ \\
$\eta$ (validity decay coefficient) & $0.8$ \\
$\omega$ (revision discount) & $0.8$ \\
$\delta_{\mathrm{stop}}$ (posterior superiority) & $0.75$ \\
$\varepsilon$ (posterior stability) & $0.02$ \\
$\gamma$ (sigmoid temperature) & $8.0$ \\
Non-Debate Agent temperature & $0.1$ \\
Type-Specific Agent temperature & $0.9$ \\
$K_{\mathrm{JERE}}$ (IERA iterations) & $1$ \\
\bottomrule
\end{tabular}
\caption{Hyperparameters of \textsc{SMADE-IE}, held fixed across the twelve evaluation settings. Symbol names follow \S\ref{sec:debate} and \S\ref{sec:alignment}.}
\label{tab:smade-hp}
\end{table}

\section{Analysis and Discussion}
\label{analysis}

\subsection{Backbone Generalization}
\label{sec:backbone-generalization}

To evaluate whether the effectiveness of \textsc{SMADE-IE} depends on a specific LLM backbone, we further conduct experiments with Gemini 3 Flash Preview, DeepSeek-V4-Pro, and Claude Sonnet 4.6. Table~\ref{tab:backbone-generalization} reports results on representative NER, RE, and JERE datasets.

\begin{table*}[t]
\centering
\small
\setlength{\tabcolsep}{11.0pt}
\begin{tabular}{llcccccc}
\toprule
\multirow{2}{*}{\textbf{Backbone}}
& \multirow{2}{*}{\textbf{Method}}
& \multicolumn{2}{c}{\textbf{OntoNotes5 (NER)}}
& \multicolumn{2}{c}{\textbf{SemEval2010 (RE)}}
& \multicolumn{2}{c}{\textbf{CoNLL04 (JERE)}} \\
\cmidrule(lr){3-4}
\cmidrule(lr){5-6}
\cmidrule(lr){7-8}
&
& $\mathrm{F1}_{\mathrm{P}}$ & $\mathrm{F1}_{\mathrm{S}}$
& $\mathrm{F1}_{\mathrm{P}}$ & $\mathrm{F1}_{\mathrm{S}}$
& $\mathrm{F1}_{\mathrm{P}}$ & $\mathrm{F1}_{\mathrm{S}}$ \\
\midrule

\multirow{5}{*}{\shortstack[l]{Gemini 3\\Flash Preview}}
& AEiO
& \underline{79.37} & \underline{68.76}
& \textbf{37.31} & \underline{21.64}
& -- & -- \\

& One-Step
& 27.23 & 20.96
& 15.39 & 4.31
& -- & -- \\

& G\&O
& 24.47 & 19.14
& 12.65 & 3.91
& -- & -- \\

& \textsc{CrossAgentIE}
& 71.95 & 60.11
& 15.84 & 4.52
& \underline{54.61} & \underline{38.85} \\

& \textbf{\textsc{SMADE-IE}}
& \textbf{88.14} & \textbf{72.95}
& \underline{30.60} & \textbf{23.73}
& \textbf{62.09} & \textbf{43.69} \\

\midrule

\multirow{5}{*}{DeepSeek-V4-Pro}
& AEiO
& 70.73 & 57.37
& \textbf{38.84} & \underline{15.38}
& -- & -- \\

& One-Step
& 29.98 & 23.68
& 19.23 & 6.76
& -- & -- \\

& G\&O
& 29.66 & 24.47
& 22.28 & 5.61
& -- & -- \\

& \textsc{CrossAgentIE}
& \underline{74.65} & \underline{63.44}
& 20.29 & 5.69
& \underline{60.35} & \underline{44.56} \\

& \textbf{\textsc{SMADE-IE}}
& \textbf{87.91} & \textbf{71.04}
& \underline{36.59} & \textbf{18.91}
& \textbf{66.03} & \textbf{53.00} \\

\midrule

\multirow{5}{*}{\shortstack[l]{Claude\\Sonnet 4.6}}
& AEiO
& \underline{78.24} & \underline{66.81}
& \textbf{35.71} & \underline{15.58}
& -- & -- \\

& One-Step
& 31.39 & 24.63
& 22.36 & 6.16
& -- & -- \\

& G\&O
& 24.94 & 19.81
& 16.79 & 6.23
& -- & -- \\

& \textsc{CrossAgentIE}
& 70.11 & 60.89
& 17.40 & 3.48
& \underline{55.61} & \underline{38.58} \\

& \textbf{\textsc{SMADE-IE}}
& \textbf{85.04} & \textbf{69.72}
& \underline{32.28} & \textbf{16.95}
& \textbf{63.65} & \textbf{44.56} \\

\bottomrule
\end{tabular}

\caption{
Backbone generalization results on representative NER, RE, and JERE datasets. Best results under each backbone are shown in \textbf{bold}, and second-best results are \underline{underlined}. ``--'' denotes unsupported settings.
}
\label{tab:backbone-generalization}
\vspace{-3mm}
\end{table*}

Across the three backbones, \textsc{SMADE-IE} consistently achieves the highest Strict F1 on all evaluated NER, RE, and JERE settings. On OntoNotes5, it also obtains the highest Partial F1 with all three backbones. For example, with DeepSeek-V4-Pro, \textsc{SMADE-IE} reaches 87.91/71.04 in $\mathrm{F1}_{\mathrm{P}}$/$\mathrm{F1}_{\mathrm{S}}$, compared with 74.65/63.44 for \textsc{CrossAgentIE}. Similar improvements are observed on CoNLL04, where \textsc{SMADE-IE} improves $\mathrm{F1}_{\mathrm{S}}$ over \textsc{CrossAgentIE} by 4.84, 8.44, and 5.98 points with Gemini 3 Flash Preview, DeepSeek-V4-Pro, and Claude Sonnet 4.6, respectively.

SemEval2010 exhibits a more nuanced pattern. AEiO achieves higher Partial F1 than \textsc{SMADE-IE} with all three backbones (37.31 vs.\ 30.60 with Gemini 3 Flash Preview, 38.84 vs.\ 36.59 with DeepSeek-V4-Pro, and 35.71 vs.\ 32.28 with Claude Sonnet 4.6), whereas \textsc{SMADE-IE} consistently achieves higher Strict F1 (23.73 vs.\ 21.64, 18.91 vs.\ 15.38, and 16.95 vs.\ 15.58, respectively). This distinction suggests that AEiO recovers more partially matched relation arguments on SemEval2010, while \textsc{SMADE-IE} produces more exact predictions under strict matching. Overall, the consistent Strict F1 gains across Gemini, DeepSeek, and Claude indicate that the benefits of sparse coordination and structured conflict resolution are not specific to a particular LLM backbone.

\begin{figure}[t]
  \centering
  \includegraphics[width=\columnwidth]{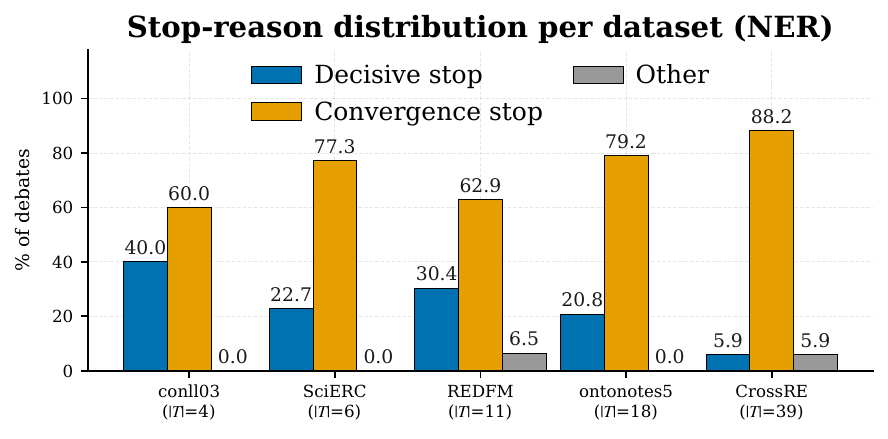}
  \vspace{0.4em}
  \includegraphics[width=\columnwidth]{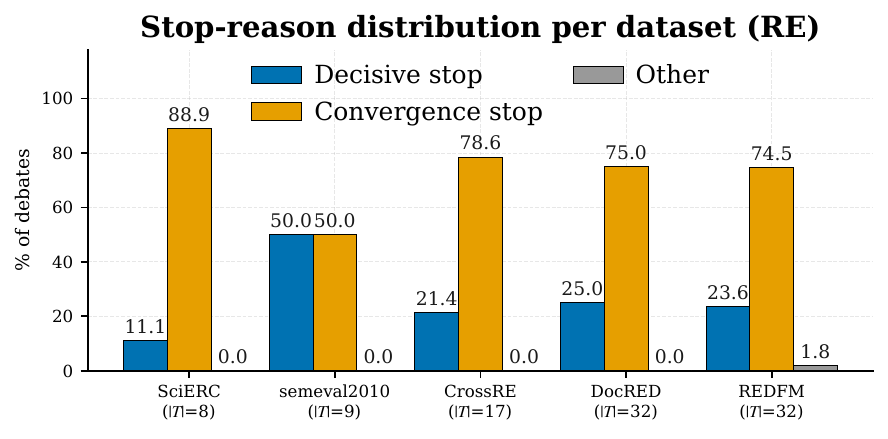}
  \caption{Early-stopping reason distributions of \textsc{SMADE-IE} on NER and RE datasets.}
  \label{fig:stop_reason}
\end{figure}

\begin{figure}[t]
  \centering
  \includegraphics[width=\columnwidth]{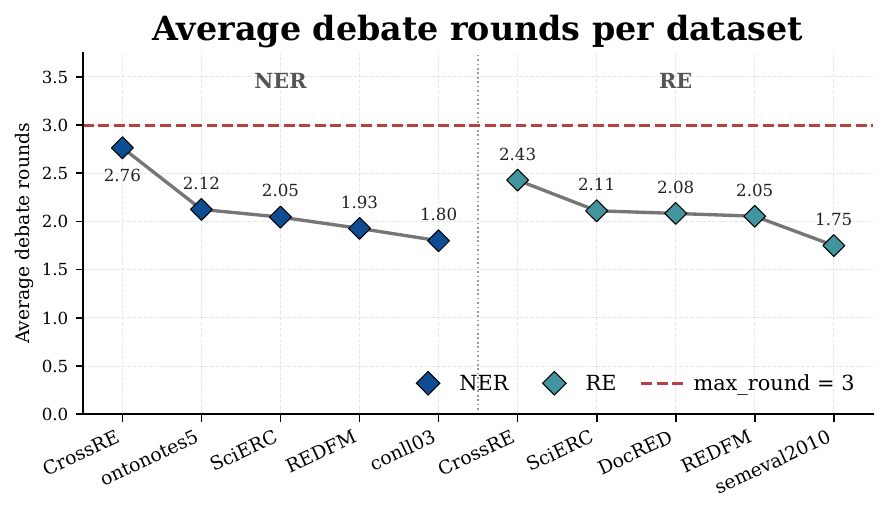}
  \caption{Average number of debate rounds under a maximum budget of 3.}
  \label{fig:stop_round}
\end{figure}

\subsection{Error Analysis}
\label{sec:error-analysis}

We conduct a fine-grained error analysis on three error categories: \emph{omission}, \emph{boundary error}, and \emph{type confusion}. An omission occurs when a gold instance has no overlapping prediction. A boundary error occurs when a prediction overlaps with the gold span and has the correct type, but does not exactly match the gold boundary. Type confusion occurs when a detected gold instance is assigned an incorrect type, regardless of whether the predicted span exactly or partially matches the gold span. Omission rates are normalized by the number of gold instances, while boundary-error and type-confusion rates are normalized by the number of detected gold instances.

\begin{table}[t]
\centering
\small
\setlength{\tabcolsep}{3.5pt}
\begin{tabular}{llcc}
\toprule
\textbf{Dataset} & \textbf{Error Type} 
& \textbf{AEiO} & \textbf{\textsc{SMADE-IE}} \\
\midrule

\multirow{3}{*}{OntoNotes5}
& Omission       & 26.0\% & \textbf{19.9\%} \\
& Boundary Error & 19.2\% & \textbf{17.2\%} \\
& Type Confusion & 20.7\% & \textbf{13.4\%} \\

\midrule
\multirow{3}{*}{REDFM (NER)}
& Omission       & 70.1\% & \textbf{22.7\%} \\
& Boundary Error & 14.0\% & \textbf{5.8\%} \\
& Type Confusion & 24.0\% & \textbf{22.7\%} \\

\midrule
\multirow{3}{*}{CrossRE}
& Omission       & 85.6\% & \textbf{63.9\%} \\
& Boundary Error & 14.4\% & \textbf{9.3\%} \\
& Type Confusion & 67.1\% & \textbf{57.5\%} \\

\midrule
\multirow{3}{*}{REDFM (RE)}
& Omission       & 74.5\% & \textbf{65.0\%} \\
& Boundary Error & 26.3\% & \textbf{14.8\%} \\
& Type Confusion & 30.8\% & \textbf{19.7\%} \\

\bottomrule
\end{tabular}
\caption{
Error analysis on NER and RE. Lower values indicate fewer errors. Omission rates are normalized by the number of gold instances, while boundary-error and type-confusion rates are normalized by the number of detected gold instances.
}
\label{tab:error-analysis}
\vspace{-3mm}
\end{table}

As shown in Table~\ref{tab:error-analysis}, \textsc{SMADE-IE} consistently reduces all three error categories relative to AEiO across the four settings. The largest reduction occurs for omissions on REDFM NER (70.1\%$\rightarrow$22.7\%). On OntoNotes5, type confusion decreases substantially (20.7\%$\rightarrow$13.4\%), while on REDFM RE, both boundary error (26.3\%$\rightarrow$14.8\%) and type confusion (30.8\%$\rightarrow$19.7\%) are markedly reduced. These results provide direct evidence that \textsc{SMADE-IE} alleviates the specific extraction errors motivating its design, rather than only improving aggregate F1 scores.

\subsection{Effectiveness of Early Stopping}
Figures~\ref{fig:stop_reason} and~\ref{fig:stop_round} report the early-stopping behavior of \textsc{SMADE-IE}.
Posterior stability is the dominant stopping reason, accounting for 60.0\%--88.2\% of early stops on NER and 74.5\%--88.9\% on most RE datasets, whereas posterior superiority is more prominent on CoNLL03 (40.0\%) and SemEval2010 (50.0\%).
The \emph{Other} category corresponds to cases in which the Ground and Warrant components of one or both debate agents are terminated.
Consistent with these distributions, \textsc{SMADE-IE} requires only 1.75--2.76 rounds on average under a maximum budget of three.
Simpler datasets such as CoNLL03 and SemEval2010 stop earliest at 1.80 and 1.75 rounds, while the more ambiguous CrossRE requires 2.76 rounds for NER and 2.43 for RE.
These results show that early stopping suppresses redundant debate while preserving additional rounds for harder cases.

\begin{figure}[t]
  \centering
  \includegraphics[width=\columnwidth]{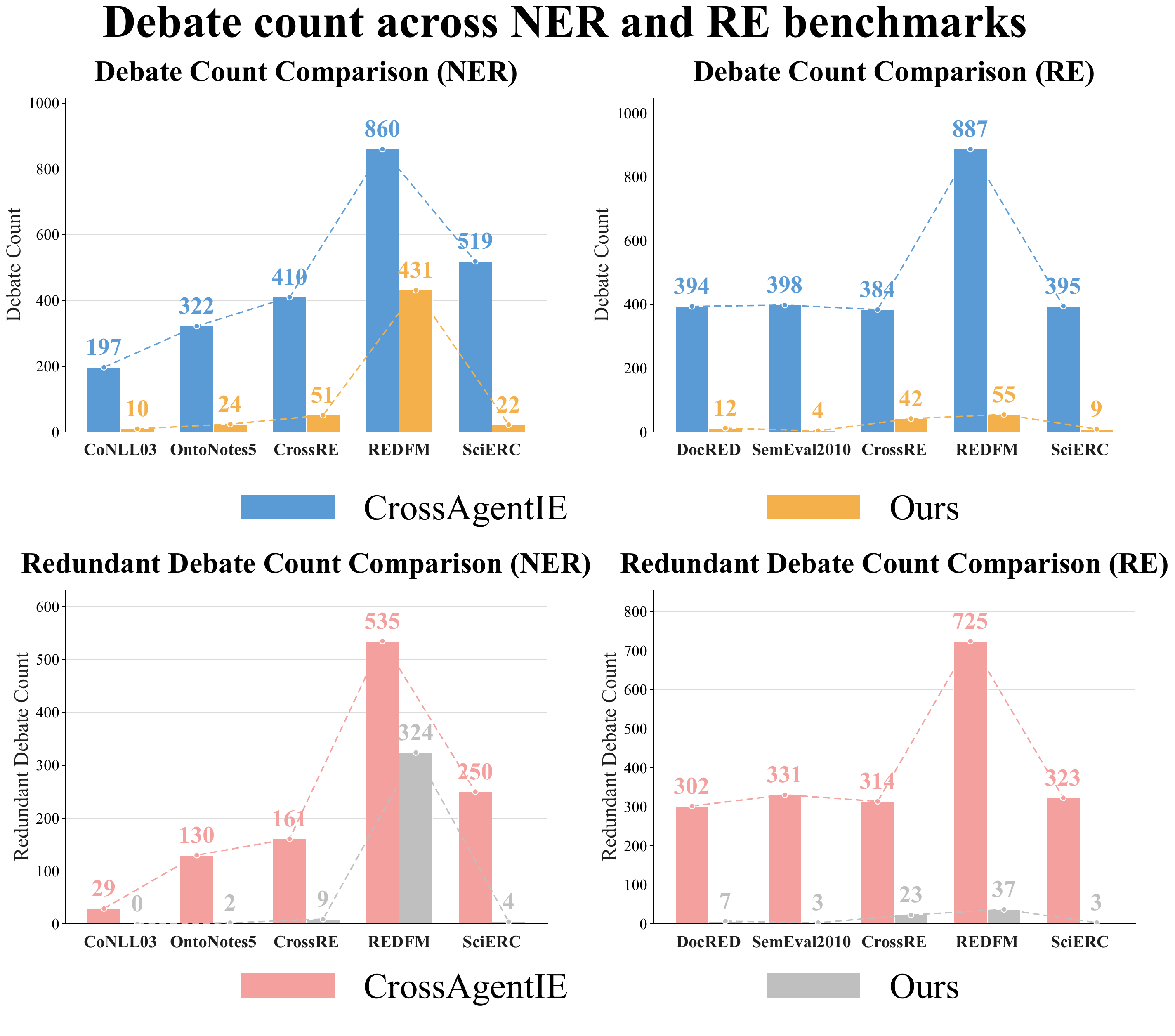}
  \caption{Debate and redundant-debate counts of \textsc{CrossAgentIE} and \textsc{SMADE-IE} on NER and RE datasets. Redundant debates refer to debates where the gold type does not exist among the proposed candidate types.}
  \label{fig:debate_efficiency}
\end{figure}

\subsection{Debate Efficiency}
Figure~\ref{fig:debate_efficiency} reports a count-level analysis of the source of token savings. Across the ten NER and RE settings, \textsc{SMADE-IE} reduces the total number of debates from 4{,}766 to 660. For NER, substantial reductions are observed on every dataset: from 197 to 10 on CoNLL03, from 322 to 24 on OntoNotes5, from 410 to 51 on CrossRE, and from 519 to 22 on SciERC. REDFM remains the most challenging setting, yet the debate count is still nearly halved, from 860 to 431. A similar trend holds for RE, where the number of debates decreases from 394 to 12 on DocRED, from 398 to 4 on SemEval2010, from 384 to 42 on CrossRE, from 887 to 55 on REDFM, and from 395 to 9 on SciERC. Redundant debates are likewise reduced from 3{,}100 to 412 overall, indicating that Router-based type selection eliminates most unnecessary comparisons rather than merely shortening informative ones.


\begin{figure*}[t]
  \centering
  \includegraphics[width=\textwidth]{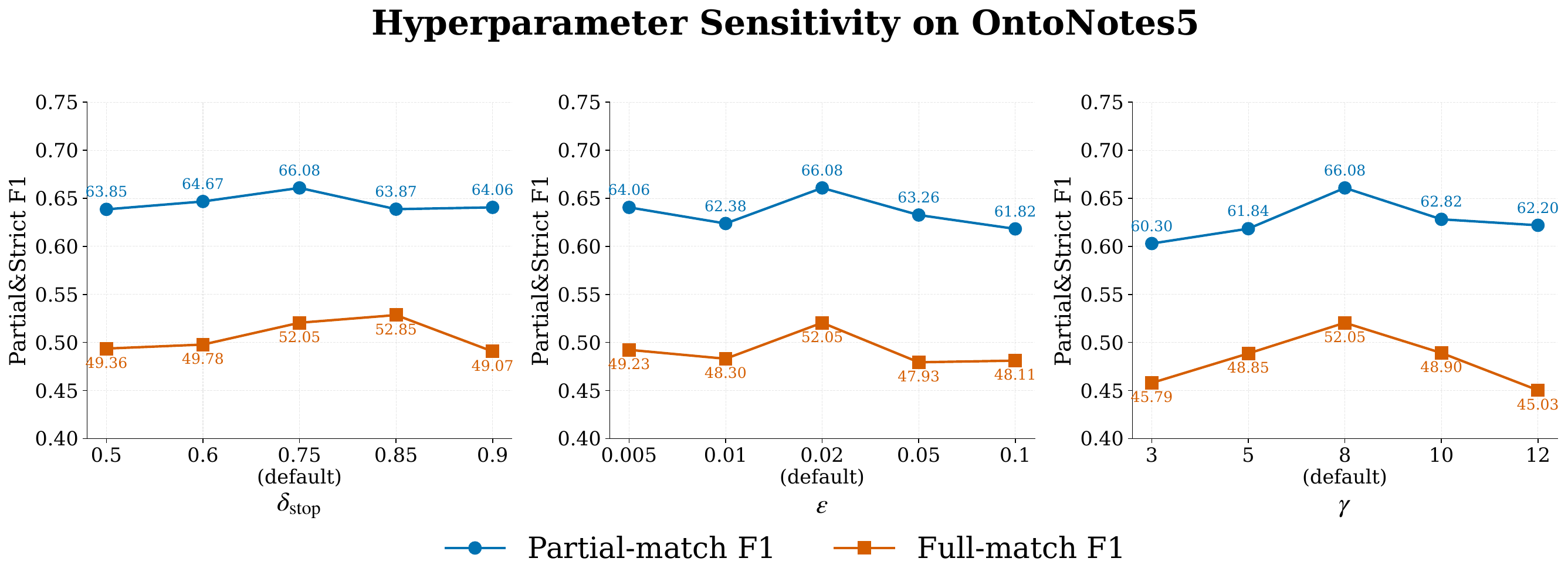}
  \caption{Hyperparameter sensitivity on OntoNotes5. We vary $\delta_{\mathrm{stop}}$, $\varepsilon$, and $\gamma$ around their default values, where the default configuration ($\delta_{\mathrm{stop}}=0.75$, $\varepsilon=0.02$, $\gamma=8$) provides the best overall trade-off.}
  \label{fig:hyperparam}
\end{figure*}

\subsection{End-to-End Inference Latency}
\label{sec:latency}

Token consumption does not fully characterize the practical overhead of multi-agent inference. We therefore measure end-to-end latency under a unified runtime environment. Experiments are conducted on a server with two NVIDIA RTX 2080 Ti GPUs (22~GB total GPU memory), 9 CPU cores of an Intel Xeon Platinum 8255C processor, and 48~GB RAM. 

\begin{table}[t]
\centering
\small
\setlength{\tabcolsep}{10.0pt}
\begin{tabular}{llcc}
\toprule
\textbf{Task} & \textbf{Dataset}
& \textbf{Mean (s)}
& \textbf{Median (s)} \\
\midrule
NER  & SciERC  & 10.6 & 10.5 \\
     & REDFM   & 26.0 & 19.0 \\
\midrule
RE   & SciERC  & 15.6 & 15.0 \\
     & REDFM   & 26.0 & 18.0 \\
\midrule
JERE & CoNLL04 & 33.7 & 33.5 \\
     & NYT     & 32.6 & 34.5 \\
\bottomrule
\end{tabular}
\caption{
End-to-end inference latency of \textsc{SMADE-IE}. Mean and median latency are reported in seconds per instance.
}
\label{tab:latency}
\vspace{-3mm}
\end{table}

As shown in Table~\ref{tab:latency}, mean latency ranges from 10.6 to 26.0 seconds per instance on NER and RE, while JERE requires approximately 33 seconds due to additional agent interactions and entity--relation alignment. Mean and median latency are similar on SciERC, CoNLL04, and NYT, whereas the larger gaps on REDFM indicate occasional slower instances. Although sparse routing reduces unnecessary agent calls, multi-agent inference still incurs non-trivial wall-clock overhead.

\subsection{Hyperparameter Sensitivity}
Figure~\ref{fig:hyperparam} reports hyperparameter sensitivity on OntoNotes5.
The default configuration provides the best overall trade-off, achieving 66.08 Partial F1 and 52.05 Strict F1.
Varying $\delta_{\mathrm{stop}}$ produces only moderate fluctuation: Partial F1 stays within 63.85--66.08 and Strict F1 within 49.07--52.85, indicating that the posterior-superiority threshold is robust.
The convergence threshold $\varepsilon$ is more influential: the default value of 0.02 outperforms both stricter and looser settings, with Partial F1 dropping to 62.38 at 0.01 and 61.82 at 0.1.
The parameter $\gamma$ exhibits the clearest trade-off.
A small value of 3 reduces F1 to 60.30/45.79, while an overly large value of 12 also hurts performance, reaching only 62.20/45.03.
These results suggest that the default configuration balances evidence accumulation and early stopping, and that \textsc{SMADE-IE} remains robust under moderate perturbations.

\section{Posterior Superiority Bound for Dual-Track Termination}
\label{app:bound}

In this section, we derive the Chebyshev-type upper bound used in \S\ref{sec:debate} and show how it motivates two complementary stopping criteria, posterior superiority and posterior stability, which jointly govern the dual-track termination mechanism.

\subsection{Setup and Notation}
Consider two competing candidates with independent Beta posteriors $p_i \sim \mathrm{Beta}(\alpha_i, \beta_i)$ for $i \in \{1,2\}$. Their posterior mean and evidence mass are
\[
\hat p_i \triangleq \frac{\alpha_i}{\alpha_i+\beta_i},
\qquad
N_i \triangleq \alpha_i+\beta_i,
\]
and the variance of $p_i$ is
\[
\sigma_i^2 \triangleq \mathrm{Var}(p_i)
=\frac{\hat p_i(1-\hat p_i)}{N_i+1}.
\]
Assume candidate $1$ currently has the larger posterior mean, i.e.,
\begin{equation}
\Delta \triangleq \hat p_1 - \hat p_2 > 0.
\end{equation}

Our goal is to bound the probability of incorrect selection, $\mathbb{P}(p_2 > p_1)$, namely the probability that the current runner-up is in fact better than the leader. This quantity is the complement of the posterior-superiority probability $\mathcal{P}(p_1 > p_2)$ used in \S\ref{sec:debate}.
\subsection{Chebyshev Bound on the Confidence Gap}
Let $D \triangleq p_1 - p_2$. Independence of $p_1$ and $p_2$ gives
\[
\mathbb{E}[D]=\Delta,
\qquad
\mathrm{Var}(D)=\sigma_1^2+\sigma_2^2.
\]
The event $\{p_2 > p_1\}$ implies $\{|D - \Delta| \geq \Delta\}$. Applying Chebyshev's inequality yields
\begin{equation}
\label{eq:cheb-general}
\small
\begin{aligned}
\mathbb{P}(p_2 > p_1)
&\leq \frac{\operatorname{Var}(D)}{\Delta^2} \\
&=
\frac{1}{\Delta^2}
\left[
\frac{\hat p_1(1-\hat p_1)}{N_1+1}
+
\frac{\hat p_2(1-\hat p_2)}{N_2+1}
\right].
\end{aligned}
\end{equation}
The bound decreases as either the posterior gap increases or more evidence is accumulated.

\subsection{Balanced-Evidence Special Case}
When the two candidates have matched evidence mass, $N_1=N_2\triangleq N$, define the average posterior mean as $\mu \triangleq (\hat p_1+\hat p_2)/2$, so that $\hat p_i = \mu \pm \Delta/2$. The numerator in Eq.~(\ref{eq:cheb-general}) then becomes
\begin{equation}
\label{eq:quad}
\begin{aligned}
f(\mu)
&=\hat p_1(1-\hat p_1)+\hat p_2(1-\hat p_2)\\
&=-2\mu^2+2\mu-\tfrac{\Delta^2}{2}.
\end{aligned}
\end{equation}
This concave quadratic attains its maximum at $\mu=1/2$, yielding the worst-case value
$f(1/2)=\frac{1-\Delta^2}{2}$.
Substituting this result into Eq.~(\ref{eq:cheb-general}) gives the symmetric bound
\begin{equation}
\label{eq:cheb-symm}
\mathbb{P}(p_2 > p_1)
\leq
\frac{1}{2(N+1)}\!\left(\frac{1}{\Delta^2}-1\right).
\end{equation}
Under balanced evidence, the probability of incorrect selection is jointly controlled by two observable quantities: the posterior mean gap $\Delta$ and the evidence mass $N$.

\subsection{Implications for Dual-Track Stopping}
Eq.~(\ref{eq:cheb-symm}) reveals two complementary regimes under which the upper bound becomes small. Each regime corresponds to one stopping signal used in \S\ref{sec:debate}.

\textbf{Posterior superiority.}
When the leading candidate is well separated from the alternatives, $\Delta$ becomes large and the bound in Eq.~(\ref{eq:cheb-symm}) decreases accordingly. The posterior-superiority criterion of Eq.~(\ref{eq:psup}) fires exactly in this regime: once $\mathcal{P}(\pi_{l^\ast}^{(t)} > \pi_{\bar{l}^\ast}^{(t)}) > \delta_{\mathrm{stop}}$, the probability of incorrect selection is sufficiently suppressed, making additional debate unlikely to overturn the current decision.

\textbf{Posterior stability.}
When candidate types are semantically similar, $\Delta$ may remain small even after multiple rounds. In this regime, the bound can only be further reduced through additional informative evidence. The validity-decay rule in Eq.~(\ref{eq:vupdate}) progressively downweights repeatedly weakened components, causing the posterior updates accumulated by Eq.~(\ref{eq:beta-update}) to diminish over time. Consequently, once the posteriors become nearly unchanged across consecutive rounds, further debate contributes little additional information. This stabilization is exactly what the squared-Hellinger criterion of Eq.~(\ref{eq:stab}) measures: terminating once $\Delta_{\mathrm{conf}}^{(t)} < \varepsilon$ matches the regime in which Eq.~(\ref{eq:cheb-symm}) cannot be tightened further without new evidence.

Together, posterior superiority addresses the regime of a dominant candidate, while posterior stability captures the regime of diminishing posterior updates. The two signals jointly control the probability of incorrect selection without requiring a fixed debate budget.

\section{Case Study}
\label{app:case}

We isolate the contribution of each \textsc{SMADE-IE} component through three contrastive cases decoded by the GPT-3.5-Turbo-0125 backbone. Each per-stage trace is contrasted with the final prediction of \textsc{CrossAgentIE} under the same backbone. The cases probe the Verification Agent on a context-dependent type mislabel in Listing~\ref{lst:case-veri}, the Review Agent and Evidence-Driven Debate on a routed-out correct type in Listing~\ref{lst:case-debate}, and the IERA loop on a joint extraction with spurious relations in Listing~\ref{lst:case-iera}.

\subsection{Case 1: Verification flips a context-dependent type mislabel}

\begin{lstlisting}[caption={OntoNotes5 NER sample 3621 trace.}, label={lst:case-veri}, captionpos=b]
[SENTENCE]
In an interview with Blender (can't remember
what issue, but Janet 's on the cover), Janet
talks at length about having mulitiple
personalities.

[GOLD]
Blender -> ORG
Janet   -> PERSON  

[SMADE-IE]
  -- NER stage --
  Router      : types = {PERSON, WORK_OF_ART},
                c = low  -> Global Extraction Mode
  Universal   : {Blender: WORK_OF_ART,
                 Janet:   PERSON}
  Verification: reject {Blender: WORK_OF_ART},
                add    {Blender: ORG}
  Final pred  = Gold                  Strict F1 = 1.00

[CrossAgentIE]
  All 18 OntoNotes types selected in parallel.
  Claims on "Blender": {PRODUCT, WORK_OF_ART}
                       -> 2-way conflict
                       (ORG agent silent).
  Debate    : free-form opinion exchange,
              summarizer tie-break -> WORK_OF_ART
              (the correct ORG label never enters
              the debate space).
  Spurious  : 'an interview' tagged as EVENT.
  Final pred = [Blender: WORK_OF_ART,
                Janet: PERSON,
                an interview: EVENT]
  Strict F1 = 0.571   
\end{lstlisting}

The Universal Agent's literal reading misclassifies \texttt{Blender} as \textsc{Work\_of\_Art}, treating the magazine title as a creative work, and would have committed to the wrong label in a single call. Because the Verification Agent receives both the candidate set and the full sentence at once, it observes the discriminating context ``interview with Blender'' and resolves the conflict by simultaneously rejecting the \textsc{Work\_of\_Art} reading and adding the \textsc{ORG} reading in one call. In contrast, \textsc{CrossAgentIE}'s all-agent claim phase admits only \textsc{Product} and \textsc{Work\_of\_Art} on this span, so the correct \textsc{ORG} label never enters the debate space; the GroupChat tie-break therefore settles on the wrong label, and the schema-wide selection additionally introduces a spurious \textsc{Event} span. Calibrated verification over a routed type subset, rather than schema-wide agent claims, enables a single additional call to recover a mistyped mention without inflating the candidate set.

\subsection{Case 2: Review recovers a routed-out type and debate confirms it}

\begin{lstlisting}[caption={REDFM NER sample 334 trace on the span \texttt{Panoramio}.}, label={lst:case-debate}, captionpos=b]
[SENTENCE]
Panoramio was a geo-located tagging, photo 
sharing mashup active between 2005 and 2016. 
Photos uploaded to the site were accessible 
as a layer in Google Earth and Google Maps. 
The site's goal was to allow Google Earth 
users to learn more about a given area by 
viewing the photos that other users had taken 
at that location. Panoramio was acquired by
Google in 2007. In 2009 the website was 
among 1000 most popular websites worldwide.

[GOLD]
Panoramio      -> media
Google         -> organization
2005, 2016, 2007 -> date

[SMADE-IE]
  -- NER stage --
  Router      : types = {concept, date, organization},
                c = medium
                # 7 of 10 REDFM types pruned
                -> Type-Centric Extraction Mode
  Type-Specific Agents on routed types:
    organization agent -> Panoramio: organization
                            Google: organization
    date         agent -> 2005, 2016, 2007: date
    concept      agent -> []
  Review Agent (covers residual 7 types):
                       -> Panoramio: media
                       # 'media' was missed by Router
                       # but recovered here.
  Conflict detected on span "Panoramio":
                       {organization, media}
                       -> promoted to debate

  -- Debate stage on "Panoramio" --
  Init q   : media = 0.448,  org = 0.364   gap = 0.084
  Round 1 pi: media = 0.623,  org = 0.449   gap = 0.174
  Round 2 pi: media = 0.693,  org = 0.418   gap = 0.275
              stop via convergence track
  Winner = media                       matches Gold

[CrossAgentIE]
  Selects all 10 REDFM types in parallel.
  Claims on "Panoramio": {organization}
                         (media agent silent
                          on this span).
  No conflict raised -> commits to organization
  directly, off-gold.
  Spurious: Google Earth / Google Maps tagged as
            'event'/'location' from distractor
            agents.
  Final pred on "Panoramio" = organization
                                       (wrong)
\end{lstlisting}

The Router selects only $\{$concept, date, organization$\}$ for this sentence, leaving the correct type \emph{media} in the residual set. The \emph{organization} agent assigns ``Panoramio'' to its type based on the plausible company-like surface form, while the Review Agent, scanning the residual seven types, also recovers ``Panoramio'' under \emph{media}. These two predictions conflict, and a binary debate between \emph{media} and \emph{organization} is launched with a narrow initial qualifier gap of $0.084$. Two rounds widen the gap to $0.275$ as the \emph{media} side cites the sentence-internal mentions of ``mashup'', ``photo sharing site'', and ``website'' as Ground, attacks the organization-side Warrant on type fit, and converges. \textsc{CrossAgentIE} selects the full ten-type alphabet upfront, yet its \emph{media} agent remains silent on this span, so no conflict is raised and the system commits to \emph{organization} without scrutiny. This case shows that Review-driven candidate recovery and structured debate are complementary: Review reinstates the type the Router would otherwise exclude, and debate then promotes the recovered candidate over the routed-but-incorrect one using evidence that a uniform multi-agent vote would not surface.

\subsection{Case 3: IERA reconciles a joint extraction with spurious relations}

\begin{lstlisting}[caption={CoNLL04 Joint sample 48 trace.}, label={lst:case-iera}, captionpos=b]
[SENTENCE]
The plane, owned by Bradley First Air, of
Ottawa, Canada, was carrying cargo to Montreal
for Emery Air Freight Corp., an air freight
courier service with a hub at the Dayton airport.

[GOLD]
(Bradley First Air, Org based in, Ottawa)
(Bradley First Air, Org based in, Canada)
(Ottawa,            Located in,    Canada)

[SMADE-IE]
  -- NER stage --
  Router      : {Loc, Org}, c = low
  Universal   : {Bradley First Air: Org,
                 Ottawa: Loc, Canada: Loc,
                 Montreal: Loc,
                 Emery Air Freight Corp.: Org,
                 Dayton airport: Loc}        (6)
  Verification: add {Dayton: Loc}             (+1)
  Entities    = 7 spans

  -- RE stage on the 7-entity set --
  Router      : {Org based in, Located in}, c = low
  Universal   :
    (Bradley First Air, Org based in, Ottawa)
    (Ottawa,            Located in,   Canada)
    (Emery Air Freight Corp., Org based in,
                              Dayton airport)
  Verification: add (Bradley First Air,
                     Org based in, Canada)
  Relations   = 4   # includes 1 spurious edge

  -- IERA pass 1 --
  RE rerun on the new entity set:
    Universal   : same 3 + (Bradley First Air,
                            Org based in, Canada)
    Verification: reject (Emery Air Freight Corp.,
                          Org based in,
                          Dayton airport)
  Relations   = 3   # spurious edge removed

  Final pred = Gold                  Strict F1 = 1.00

[CrossAgentIE]
  7 relations emitted:
    (Bradley First Air, Org based in, Ottawa)
    (Emery Air Freight Corp., Org based in, Montreal)
    (Emery Air Freight Corp., Org based in, Dayton)
    (Ottawa,            Located in,   Canada)
    (Montreal,          Located in,   Canada)
    (Dayton,            Located in,   Canada)
    (Dayton,            Located in,   Dayton airport)
  Strict F1 = 0.40  
\end{lstlisting}

Single-pass joint extraction over-emits relations on co-mentioned organizations and locations: the RE Universal Agent interprets ``hub at the Dayton airport'' as an \textsc{Organization based in} edge, while the NER step extracts every location-like span. IERA re-invokes RE under the ontology-reconciled entity set, at which point the Verification Agent rejects the spurious \textsc{Emery--Dayton airport} edge that no longer has supporting evidence, and the loop reaches a fixed point with three correct relations. \textsc{CrossAgentIE} lacks this consistency loop and emits five additional locative edges among the same co-mentioned spans. This loop is the mechanism that converts locally plausible candidates into globally consistent triples.

\paragraph{Summary.} The three traces share a single mechanism: \textsc{SMADE-IE} allocates computation in proportion to instance difficulty. One additional call resolves a context-dependent type misclassification, Review combined with debate rescues a routed-out type, and a short loop reconciles the joint extraction. \textsc{CrossAgentIE} applies a uniform all-type, all-debate budget and incurs noise from every distractor type it admits. Calibrated allocation, rather than larger ensembles, is what translates agent collaboration into reliable predictions.

\section{Prompts}
\label{app:prompts}

In this section, we list the prompts used by every agent in \textsc{SMADE-IE}. System prompts are fixed; user prompts are instantiated by filling the angle-bracketed fields with instance-specific values. All agent outputs are returned as JSON and parsed with tolerance for minor formatting inconsistencies; on parse failure, the pipeline reverts to the safe default of the corresponding agent in \S\ref{sec:method}. Dataset-specific type definitions are loaded from each dataset's ontology at runtime and are therefore elided from the templates below.

\subsection{Router Agent Prompt}
As formalized in Eq.~\eqref{eq:selector}, the Router Agent maps a sentence $s$ and the entity type set $\mathcal{T}_e$ to a selected subset $\mathcal{A} \subseteq \mathcal{T}_e$ together with a complexity label $c \in \{\mathrm{low}, \mathrm{med}, \mathrm{high}\}$. To preserve high recall at minimal overhead, the agent performs lightweight type selection without per-type rationales and consumes a single LLM call per sentence.

\begin{lstlisting}
SYSTEM
You are a Router Agent for Entity Extraction.
Scan the text and identify which types are present.
Type Definitions:
{type_definitions}

Classify the input as low, medium, or high.
low:    explicit, unambiguous items with few clearly separated types.
medium: multiple types, mild ambiguity, fuzzy boundaries, or moderate context dependence.
high:   strong ambiguity, overlapping candidates, implicit context, or many interacting types.

For hierarchical overlaps, choose the most specific type.
List each type at most once regardless of how many instances of it appear in the text.

USER
Identify which types are present in the text.
Text: {sentence}

JSON: {"possible_types": ["<type>", ...],
       "complexity": "<low|medium|high>"}
\end{lstlisting}

For the RE instantiation, ``Entity Extraction'' is replaced by ``Relation Extraction'' and the type definitions are drawn from the relation ontology.

\subsection{Universal Agent Prompt}
Eq.~\eqref{eq:universal} specifies the Universal Agent, which serves the Global Extraction Mode on low-complexity inputs and extracts, in a single LLM call, all candidates whose types lie in $\mathcal{T}_e$.

\begin{lstlisting}
SYSTEM
You are a Universal Entity Extraction Agent covering all types.
Extract the exact noun phrase as the span. Do not include determiners (the, a, an) or unnecessary adjectives.
An entity can only belong to one category without overlap.

Type Definitions:
{type_definitions}

USER
Extract all entities whose type is one of: {selected_types}. Return empty dict if none.

Text: {sentence}

JSON: {"<verbatim span>": "<type>", ...}
\end{lstlisting}

For RE, the JSON template becomes \texttt{\{"<head>|||<tail>": "<type>", ...\}}.

\subsection{Review Agent Prompt}
Eq.~\eqref{eq:review} defines the Review Agent, which operates under the Type-Centric Extraction Mode on medium- and high-complexity inputs. It covers the residual type set $\mathcal{A}^* = \mathcal{T}_e \setminus \mathcal{A}$, recovering candidates whose types lie outside the Router-selected subset and compensating for any Router omission. The Review Agent reuses the system prompt of the Universal Agent and adopts the user prompt below, in which \texttt{remaining\_types} instantiates $\mathcal{A}^*$.

\begin{lstlisting}
USER
Extract entities whose type is only in {remaining_types}. Assign exactly one type per entity; return [] if none.

Text: {sentence}

JSON: {"results": [{"span": "<verbatim text>",
                    "type": "<type>"}, ...]}
\end{lstlisting}

For RE, the JSON template replaces \texttt{"span"} with the two fields \texttt{"head"} and \texttt{"tail"}.

\subsection{Verification Agent Prompt}

Per Eq.~\eqref{eq:verify}, the Verification Agent is invoked only on the low-complexity path, after Universal Agent extraction. Given the candidate set, it removes incorrectly typed candidates and inserts missed entities in a single LLM call, improving extraction quality while keeping the entire path within two calls.

\begin{lstlisting}
SYSTEM
You are a Verification Agent for Entity type verifications.
Extract the exact noun phrase as the span. Do not include determiners (the, a, an) or unnecessary adjectives.
If a candidate span contains multiple independent entities separated by punctuation, reject the combined span and add each individual entity separately.
Type Definitions:
{type_definitions}

USER
Text: "{sentence}"

Candidates:
"{span_1}" is {type_1}
"{span_2}" is {type_2}
...

Reject candidates with incorrect types and add any missing entities from the text. The type must be one of the defined types.

JSON: {"added":    [{"span": "<verbatim text>", "type": "<type>"}],
       "rejected": [{"span": "<verbatim text>", "type": "<type>"}]}
\end{lstlisting}

For RE, each candidate is rendered as a triple \texttt{(<head>, <type>, <tail>)}, and the JSON fields replace \texttt{"span"} with \texttt{"head"} and \texttt{"tail"}.

\subsection{Type-Specific Agent and Toulmin-Guided Argument Construction Prompt}
Each Type-Specific Agent is bound to a single candidate type $t_\mathcal{A}^i \in \mathcal{A}$ and operates in two phases. The extraction phase, governed by Eq.~\eqref{eq:type-agent}, scans the sentence for spans of the bound type and contributes them to the candidate set $E_\mathcal{A}$ of the Type-Centric Extraction Mode. When a candidate entity $\tilde{e}^i$ is later contested by multiple types $\{\tilde{t}^i_1, \dots, \tilde{t}^i_N\}$, the agent for type $\tilde{t}^i_n$ enters the debate phase and emits the Toulmin argument $\mathcal{D}_n = \{C_n, G_n, W_n, B_n, R_n, Q_n\}$ specified by Eq.~\eqref{eq:toulmin}. The qualifier $Q_n$ is not produced by the agent itself but is computed externally by the scorer $\phi$ over the argument paragraph; the rebuttal $R_l$ of each surviving candidate $l$ is reused pre-debate to set its prior strength $\kappa_l$ via the rebuttal-honesty derivation of Eq.~\eqref{eq:kappa}.

The extraction phase uses the per-type system and user prompts:

\begin{lstlisting}
SYSTEM
You are an Entity extraction Agent for type: {agent_type}.
Definition of {agent_type}: {type_definition}
Extract the exact noun phrase as the span. Do not include determiners (the, a, an) or unnecessary adjectives.
An entity can only belong to one category without overlap.

USER
Extract all "{agent_type}" entities from the text. Return empty list if none.

Text: {sentence}

JSON: ["<span1>", "<span2>", ...]
\end{lstlisting}

In the debate phase, the same agent is reinstantiated under a debate-specific system message that supplies the contested span, the sentence, and all competing type definitions. The Toulmin-generation user prompt is then issued once per conflict:

\begin{lstlisting}
SYSTEM
You are {agent_type} Agent in a debate over the entity "{span}".
Sentence: {sentence}

Type definitions:
- {type_1}: {def_1}
- {type_2}: {def_2}
...

In each round you receive an opponent's textual evidence Ground and type-fit reasoning Warrant. Write counters that both support {agent_type} and rebut the opponent's claim. If your own components get attacked enough, you will be asked to revise them.

USER
Conflict detected on span: "{span}" in "{sentence}"

Construct a Toulmin argument as JSON:
{
  "Claim":    "The correct type for {span} is {agent_type}",
  "Ground":   "<factual evidence quoted from the text>",
  "Warrant":  "<reasoning rule linking Ground to Claim>",
  "Backing":  "<additional linguistic or domain support>",
  "Rebuttal": "<one concise sentence that presents a counter-view conflicting with the Claim, quoting an original sentence fragment as evidence; no conditionals, no hedging>"
}
\end{lstlisting}

\subsection{Refutation Prompt}
In each debate round, the opposing agent $\mathcal{D}_{\bar{l}}$ attacks the defended agent $\mathcal{D}_{l}$ over the active subset $\mathcal{V}_l^{(t)} \subseteq \{G, W\}$ defined by Eq.~\eqref{eq:active-set}. A single LLM call yields one refutation $u_{\bar{l},m}$ per defended component $m \in \mathcal{V}_l^{(t)}$; each refutation is scored against the defender by the evidence scorer $\phi$ and then folded into the Beta-posterior update of Eq.~\eqref{eq:beta-update}.

\begin{lstlisting}
USER
Counter the {target_type} Agent's argument.
Ground:  cite a fragment from the text that supports {attacker_type} and shows the opponent's evidence does not establish {target_type}.
Warrant: argue from the type definitions that {attacker_type} fits the entity and {target_type} does not.
Take a new angle if you have replied earlier.

Opponent's Ground:  {ground_text}
Opponent's Warrant: {warrant_text}

JSON: {"Ground":  "<counter-claim>",
       "Warrant": "<counter-claim>"}
\end{lstlisting}

When only Ground or only Warrant remains in $\mathcal{V}_l^{(t)}$, the instruction line, opponent excerpt, and JSON field corresponding to the inactive component are omitted; the per-round LLM budget therefore scales with the number of attackable components rather than with their nominal count.

\subsection{Revision Prompt}
When the decay rule of Eq.~\eqref{eq:vupdate} drives one or more components of an agent's argument to a validity at or below the activity threshold $\theta_v$, the owning agent must revise them in the next round. All revisions belonging to the same agent are batched into a single LLM call, so the per-round revision cost scales with the number of agents rather than with the total number of components to revise.

\begin{lstlisting}
USER
Current Toulmin argument:
Claim:    "..."
Ground:   "..."
Warrant:  "..."
Backing:  "..."
Rebuttal: "..."

Your {forfeited_components} were attacked. Revise with stronger arguments.

JSON: {"<comp_1>": "<revised text>",
       "<comp_2>": "<revised text>"}
\end{lstlisting}

The JSON output contains exactly the components listed in \texttt{forfeited\_components}; non-forfeited components are not regenerated.

\subsection{Iterative Entity--Relation Alignment Prompt}
As described in \S\ref{sec:alignment}, the Consistency Agent is invoked whenever a relation in $\bar{R}^{(t)}$ violates the ontology constraints with respect to the entity types in $\bar{E}^{(t+1)}$. All conflicts detected within the same sentence are batched into a single LLM call. For each conflict, the agent returns one of two decisions. Under \texttt{NER\_correct}, the offending relation is added to the blacklist $B^{(t)}$ and removed from $\bar{R}^{(t)}$. Under \texttt{RE\_correct}, the conflicting entity--type assignment is included in $C_-^{(t)}$, while the constraint-compatible assignment returned by the agent is included in $C_+^{(t)}$, corresponding to the entity-type correction defined in \S\ref{sec:alignment}.

\begin{lstlisting}
SYSTEM
You are a consistency arbitrator for NER--RE conflicts.
When an entity's type contradicts a relation's type constraint, decide which side is correct.

Entity Ontology:
{entity_definitions}

Relation Ontology:
{relation_definitions}

USER
Text: "{sentence}"

You have {N} conflict(s) to resolve:

Conflict 1:
  Entity:   "{entity_span_1}" typed as "{entity_type_1}" by NER
  Relation: ({head_span_1}, {relation_type_1}, {tail_span_1})
  Role:     {role_1}
  Expected types: [{expected_types_1}]
...

For each conflict, decide:
- "NER_correct": the entity type is correct; remove the relation.
- "RE_correct":  the relation is correct; replace the entity type with one of the expected types.

JSON: {"decisions": [{"conflict_id":    <1-based index>,
                      "decision":       "<NER_correct|RE_correct>",
                      "corrected_type": "<type if RE_correct, else empty>"}, ...]}
\end{lstlisting}

\subsection{Span Classification for \texorpdfstring{RE\,$\rightarrow$\,NER}{RE -> NER} Back-Fill Prompt}
The span-classification step of \S\ref{sec:alignment} assigns an entity type to every span that appears as a head or tail in $R$ but is absent from $E$. The Universal Agent is repurposed in span-classification mode with the user prompt below; its system prompt is identical to the one used by the Universal Agent elsewhere in this appendix.

\begin{lstlisting}
USER
Please classify the entity type for each span below. Use only the defined types.

Text: {sentence}
Spans to classify: ["{span_1}", "{span_2}", ...]

JSON: {"classifications": [{"span": "<verbatim text>",
                            "type": "<type>"}, ...]}
\end{lstlisting}

\end{document}